\documentclass[10pt,twocolumn,letterpaper]{article}

\usepackage{wacv}
\usepackage{times}
\usepackage{epsfig}
\usepackage{graphicx}
\usepackage{amsmath}
\usepackage{amssymb}

\usepackage{epsfig}
\usepackage{graphicx}
\usepackage{amsmath}
\usepackage{amssymb}
\usepackage{multirow}
\usepackage{graphics}
\usepackage{graphicx,booktabs,array}

\usepackage{amsmath,amsfonts,amssymb,epsfig,verbatim,color}
\usepackage{multirow}
\usepackage{rotating}
\usepackage{booktabs}
\usepackage{pdfpages}
\usepackage{amsthm} 
\usepackage{subfig}
\usepackage{algorithm}
\usepackage{algpseudocode}
\usepackage{url}
\usepackage{bm}
\usepackage{array}
\usepackage{xcolor,colortbl}
\usepackage{textcomp}

\newcolumntype{d}[1]{D{.}{.}{#1}}

\newcommand{\argmin}{\operatornamewithlimits{\arg\,\min}}

\makeatletter
\newcommand{\removelatexerror}{\let\@latex@error\@gobble}
\makeatother
\DeclareMathAlphabet{\mathpzc}{OT1}{pzc}{m}{it}
\DeclareMathAlphabet{\mathpzc}{OT1}{pzc}{m}{it}

\wacvfinalcopy 

\def\wacvPaperID{526} 

\ifwacvfinal\fi
\begin{document}

\title{EgoReID Dataset: Person Re-identification in Videos Acquired by Mobile Devices with First-Person Point-of-View}

\author{
\begin{tabular}{ccc}
Emrah Basaran\textsuperscript{1*} &
Yonatan Tariku Tesfaye\textsuperscript{2*} &
Mubarak Shah\textsuperscript{2}\\
{\tt\small basaranemrah@itu.edu.tr} &
{\tt\small yonatantariku@knights.ucf.edu} &
{\tt\small shah@crcv.ucf.edu}
\end{tabular}\\\\
\begin{tabular}{c}
\textsuperscript{1}Dept. of Computer Engineering, Istanbul Technical University\\
\textsuperscript{2}Center for Research in Computer Vision, University of Central Florida\\
\end{tabular}
}


\maketitle
\ifwacvfinal\thispagestyle{empty}\fi


\maketitle

\begin{abstract}
In recent years, we have seen the performance of video-based person Re-Identification (ReID) methods have improved considerably. However, most of the  work in this area has dealt with videos acquired by {\em fixed} cameras with wider field of view. Recently, widespread use of wearable cameras and recording devices such as cellphones have opened the door to  interesting research in first-person Point-of-view (POV) videos (egocentric videos). Nonetheless, analysis of such videos is challenging 
due to  factors such as poor video quality due to ego-motion, blurriness, severe changes in lighting conditions and perspective distortions. To facilitate the research towards conquering these challenges, this paper contributes a new dataset called EgoReID. The dataset is captured using 3 mobile cellphones with non-overlapping field-of-view. 
It contains 900 IDs and around 10,200 tracks with a total of 176,000 detections. The dataset also contains 12-sensor meta data e.g. camera orientation pitch and rotation  for each video. \\

In addition, we propose a new framework which takes advantage of both visual and sensor meta data to successfully perform Person ReID. We  extend image-based re-ID method employing  human body parsing trained on ten datasets to  video-based re-ID. In our method, first  frame level local features are extracted for each semantic region, then  3D convolutions are applied to encode the temporal information in each sequence of semantic regions. 
Additionally, we  employ sensor meta data to predict targets' next camera and their estimated time of arrival, 
which considerably improves our ReID performance as it significantly reduces our search space.

\end{abstract}
\let\thefootnote\relax\footnote{\textsuperscript{*}Authors contributed equally}
 

\begin{figure}[t]
\begin{center}
 \includegraphics[width=1.0\linewidth,height=6.3cm]{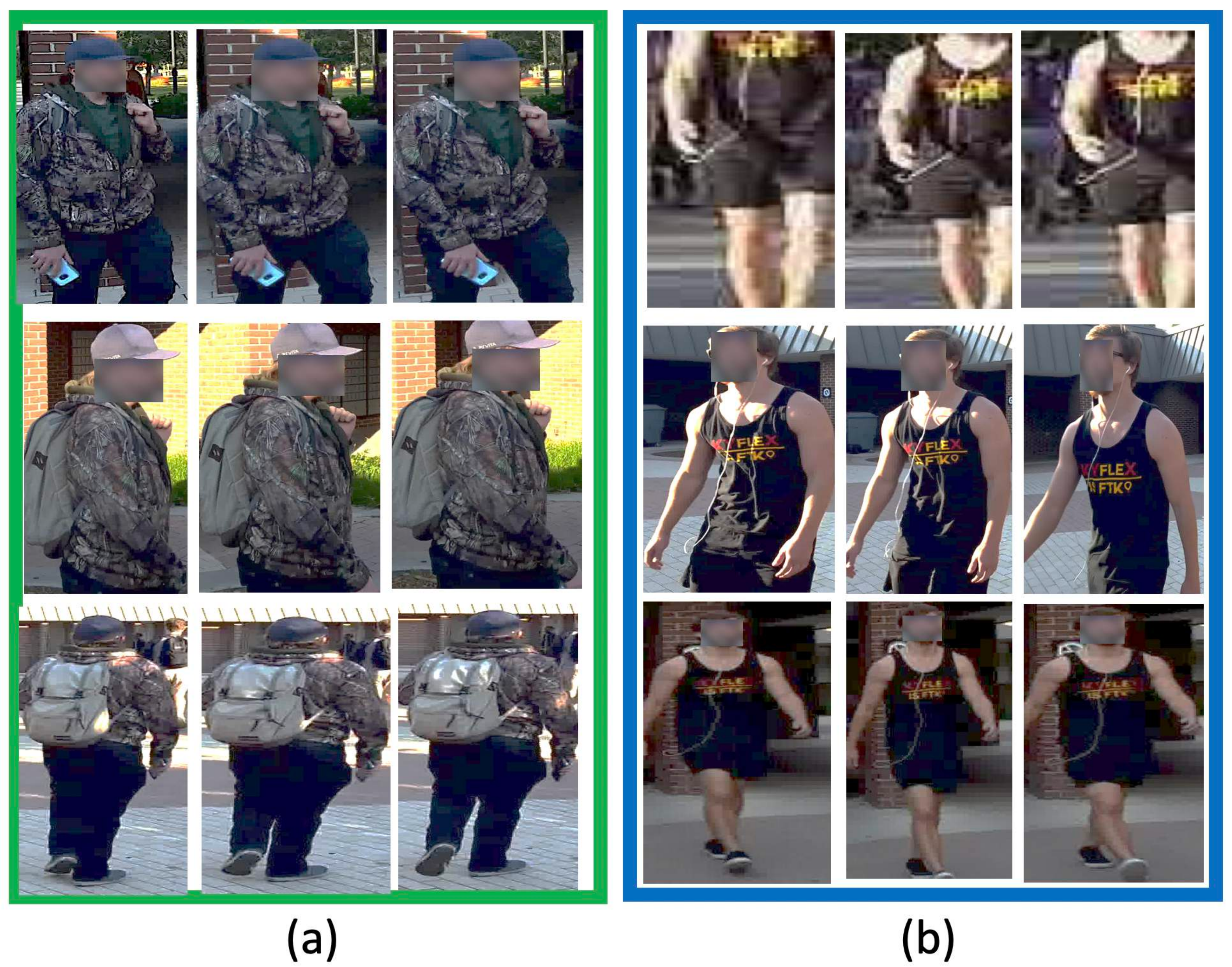}
\end{center}
  \caption{(a) shows three tracklets of the same identity from the same camera, its evident to see that due to both camera and target motions the  target is captured in different poses, backgrounds and illuminations. (b) each row shows tracklets of the same identity captured from camera 1, 2 and 3, respectively. Due to close proximity of target and the cameras, different body regions of the target are missing in Camera 1 and 2. 
  }
 
\label{fig:Tracklet_eg}
\end{figure}

\vspace{-6mm}

\section{Introduction}


Person Re-Identification (ReID) aims at associating the same pedestrian across multiple cameras \cite{gong2014person, zheng2016person}. This task has drawn increasing attention in recent years due to its importance in applications, such as surveillance \cite{wang2013}, activity analysis \cite{loy2009} and tracking \cite{yu2013harry}. Person ReID remains a challenging problem because of complex variations in camera viewpoints, human poses, lighting, occlusions, and background clutter.

Currently, in almost all attempts to solve person ReID problem, the source of data is from {\em fixed} cameras with wider field of view (FOV). Mostly, in such datasets, full bodies of pedestrians are captured and intra camera variations of  targets are very small, as pose of targets often do not change  within the same camera.
On the contrary, videos from a moving first-person 
point-of-view (POV) devices pose  unique challenges, such as frequent occlusion of targets due to close proximity of camera and targets, blurriness caused by camera motion, severe lighting and background changes, both within and across cameras,  appearance differences due to   different perspectives. Moreover, due to constant motion of both target and cameras, there are  frequent inter camera appearance changes between sequences of the same target. This is mainly due to the fact that a target can appear in several poses in a single camera view as shown in Fig. $\ref{fig:Tracklet_eg}$ (a). 

Besides the unique challenges posed by the nature of the video, these devices (such as cellphones) 
provide additional source of information. Most new generation devices are equipped with additional sensors such as gyros, accelerometers, magnetometers, GPS and are Internet-enabled, it is now possible to obtain large amounts of first-person point-of-view (POV) data un-intrusively. 


In order to facilitate the research towards such unique video domain, 
In this paper, we propose a new dataset called EgoReID. 
Different from existing datasets EgoReID has  several new features. 1)  3 synchronized mobile cellphones with non-overlapping FOV are used to record the videos. 2) Throughout the recording, all 3 cameras are moving around campus covering larger area, resulting in complex scene transformation and background. 3) 12 sensor meta data along with the videos are also collected. 4) YOLO9000 \cite{redmon2017yolo9000} and FCDSC \cite{Tesfaye2019} are used  for pedestrian detection and tracklet generation, respectively, which will be provided with the dataset. 

Besides the new dataset, we propose a novel approach to leverage network trained on 10 {\em image-based} ReID datasets for {\em video-based} Person Re-ID problem. Here, we employ a model, trained on these  image-based datasets, to extract frame level local features for each semantic region, then we employ 3D convolutions to encode the temporal information in each sequence of regions. 
 Instead of directly encoding the whole sequence, we extract features from each body regions to learn discriminative local features. We then scale each local feature with their corresponding learned weights.

In addition to traditional video features,  we successfully demonstrate the use of  sensor {\em meta data}  to complement appearance cues. In particular, we use meta data information to estimate target's next camera and its time of arrival,  that way  we significantly reduce the search space by pruning several unreliable matches which violate the estimated time of arrival constraint. 

The contributions of this paper can be summarized as follows: 
\begin{itemize}
    \item A unique and challenging EgoReID dataset is presented  and will be publicly released with detections, tracks and sensor meta data. 
    \item We propose a new model to solve person ReID problem in egocentric videos, where per frame human semantic parsing is employed to encode local visual cues. Then we extract  video feature representations by aggregating the temporal information within each sequence of body parts. 
    \item We propose a method to employ sensor meta data information to significantly reduce our search space by first predicting the next camera where the  target may appear, then estimating its time of arrival. 
    
\end{itemize}

\section{Related Works}

In this section, we review related works in the areas of video-based person re-identification and egocentric vision.

\noindent\textbf{Video-based person re-identification:}
The success of deep learning in wide range of computer vision areas has been inspiring a lot of studies in person re-identification.
In recent years, researchers are considering more realistic scenarios such as larger dataset \cite{zheng2015scalable, zheng2016}, complex scenarios \cite{zheng2017person,ristani2016performance} and  combining different modalities such as text descriptions in their approaches \cite{zheng2017unlabeled,li2017person}.
The effectiveness of Convolutional Neural Network (CNN) in learning discriminative image representations from large scale datasets have been exploited effectively in \cite{varior2016gated, wang2016joint, liu2016spatio,li2017learning,chen2017beyond, zhao2017spindle,zhou2017see}. Utilization of video data is further facilitated by the powerful feature learning ability of Convolutional Neural Networks. In video-based ReID \cite{mclaughlin2016recurrent, zhou2017see, you2016top,wang2016person, zhu2018video, ma2017person}, the learning algorithm is given a pair of video sequences instead of images. 
Authors in \cite{mclaughlin2016recurrent} introduce an RNN model to encode temporal information. The features from all timesteps are then combined using temporal pooling to compute an overall appearance feature for the complete sequence and then  the feature similarity of two videos is computed. In \cite{wang2016person}   the most discriminative video fragments are selected from noisy/incomplete image sequences of people from which reliable space-time and appearance features are computed, whilst simultaneously learning a video ranking function for person ReID. Authors in \cite{ma2017person}, introduce a new space-time person representation by encoding multiple granularities of spatio-temporal dynamics in form of time series. 
In \cite{zhou2017see}  attention weights are combined per-frame with visual features and the forward propagated RNN hidden variables, in order to weaken the influence of the noisy samples. 
Authors in \cite{xu2017jointly} employ spatial pooling layer to select regions from each frame, while temporal attention pooling is performed to select informative frames in the sequence. 
In \cite{si2018dual}, the dual attention mechanism is used, in which both intra-sequence and inter-sequence attention strategies are employed for feature refinement and feature-pair alignment, respectively. 

Unlike most of the above approaches, our  proposed method assigns weights to each local feature extracted from different body regions based on their discriminative abilities. Due to lack of large number of re-ID video datasets, we leverage image-based re-ID model trained on 10 datasets and encode temporal information in each body region by applying 3D convolutions.  Moreover, we employ sensor meta-data to further improve our ReID performance.

\begin{table*}[h!]
\centering
\begin{tabular}{|c| c| c| c |c|c|} 
 \hline
\multicolumn{1}{|c|}{}& \multicolumn{3}{|c|}{Fixed Cameras} &  \multicolumn{2}{|c|}{Mobile Cameras} \\
 \hline
 Datasets & MARS&	iLIDS&	PRID&Fergani et al.	\cite{fergnani2016body} &{\bf EgoReID}
(Ours)  \\ [0.5ex] 
 \hline 
 \hline
Number of IDs&	1,261	&300&	200	&8	&900 \\\hline
 Tracklets \#	&20,478	&600&	400	&-&	10,200\\\hline
 Camera \# &	6&	2	&2	&1&	3\\\hline
  Generated by	& DPM\cite{felzenszwalb2009object}+GMMCP \cite{dehghan2015gmmcp} &	hand&	hand&	DPM \cite{felzenszwalb2009object}	&Yolo9000 \cite{redmon2017yolo9000}+FCDSC \cite{Tesfaye2019}\\\hline
Evaluation	&mAP + CMC&	CMC&	 CMC&	mAP	&mAP + CMC \\\hline
Mode & Video-based& Video-based &Video-based &Image-based &Video-based \\\hline
Meta-data & -&-&-&-& 12 sensor meta-data \\ [1ex] 
 \hline
\end{tabular}
\caption{Summary of datasets acquired by both fixed and mobile cameras. }

\label{table:Datasets}
\end{table*}

\noindent\textbf{Egocentric vision:}
Recently, visual analysis of egocentric videos has been a hot topic in computer vision, and the work ranging from object detection \cite{fathi2011learning} to recognizing daily activities \cite{fathi2012learning, fathi2011understanding}, predicting gaze behavior \cite{li2013learning, borji2014look}, and video summarization \cite{lu2013story} have been proposed. In \cite{ferland2009egocentric}, depth perception and 3D reconstruction are improved by fusing information from egocentric and exocentric vision with laser range data. Moreover, the relationship between moving and static cameras have been studied in some works. Authors in \cite{alahi2008master, alahi2008object} improve object detection accuracy by exploring the relationship between mobile and static cameras. 
Authors in \cite{bettadapura2015egocentric} localize egocentric field-of-views using first-person point-of-view devices by matching images and video with the reference corpus, and refine the results using the first-person’s head orientation information, obtained using the  sensor devices. 

Authors in \cite{ardeshir2016ego2top} propose a framework for the identification of egocentric viewers, by matching and assigning the viewers among the egocentric videos and a top view video. In \cite{yao2019unsupervised}, authors introduced a new egocentric dataset for the traffic accident detection and proposed an unsupervised framework to detect the accidents from egocentric videos. In \cite{aghaei2016multi}, multiple face tracking problem is addressed employing the sequences of egocentric videos. Using egocentric videos captured by single camera, authors in \cite{jiang2017seeing} propose a method to infer 3D full body pose. Authors in \cite{fergnani2016body}, collected a ReID dataset using the front camera of an eye tracking device. They collected images of 8 persons with around 100 images  each.

To the best of our knowledge no work has been reported on video-based person ReID in egocentric videos. This probably is due to the lack of such a dataset. Therefore, proposed work will fill in an important gap in the literature.


\section{EgoReID Dataset}
In this section, we introduce our new EgoReID dataset.


\subsection {Dataset Description}
 
\noindent\textbf{Data collection:} We recorded the videos using 3 synchronized Samsung mobile cellphones with non-overlapping field of views. Throughout the period of recording, all 3 cameras were constantly moving around and covered larger area. The videos are recorded for $\sim20$ minutes with 30 fps. \\

 \noindent\textbf{Data preparation:} We first detect pedestrians from the videos using Yolo9000 person detector \cite{redmon2017yolo9000}. We manually verify the detection results by removing bounding boxes which do not contain a person. Then, we employ FCDSC \cite{Tesfaye2019} tracker to generate tracks. Similarly, here also we manually verify tracks generated by FCDSC using the following criteria:  1) we remove tracks which are shorter than 16 frames. 2) tracks which do not contain the target for more than 80\% of the track (containing $>$20\% false positive) are removed. 3) the detection bounding box from a track is removed if target is occluded for more than 80\%. During labeling, we manually assign the same label for track of the same person in different cameras. Three people were involved during both data collection and preparation. Each person has spent around 160 hours to finalize the dataset generation.  

 
\begin{figure*}[t]%
    \centering
    \subfloat[]{{\includegraphics[height=4.cm,width=7.5cm]{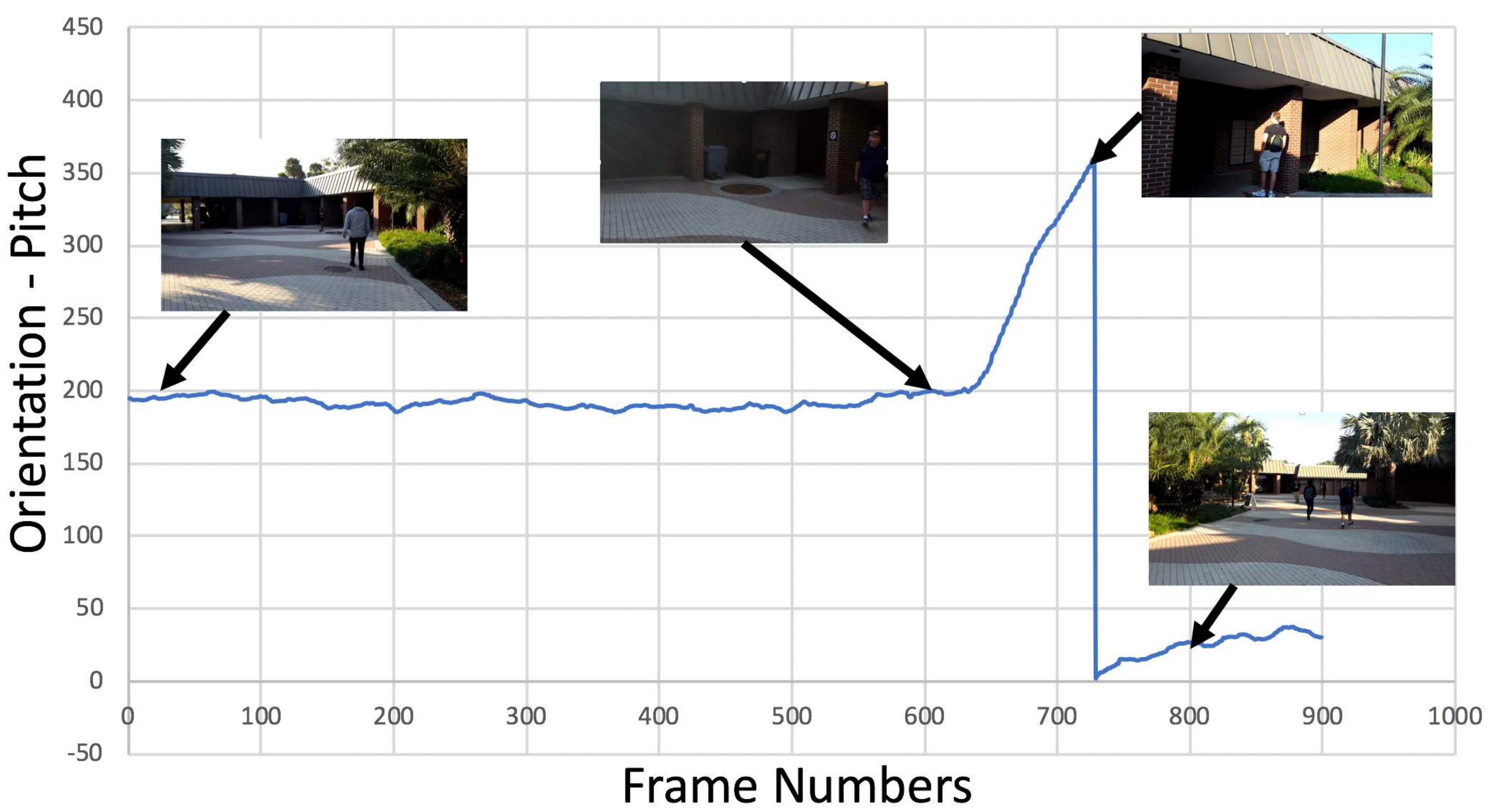}}}%
    \qquad
     \subfloat[]{{\includegraphics[height=4.cm,width=7.5cm]{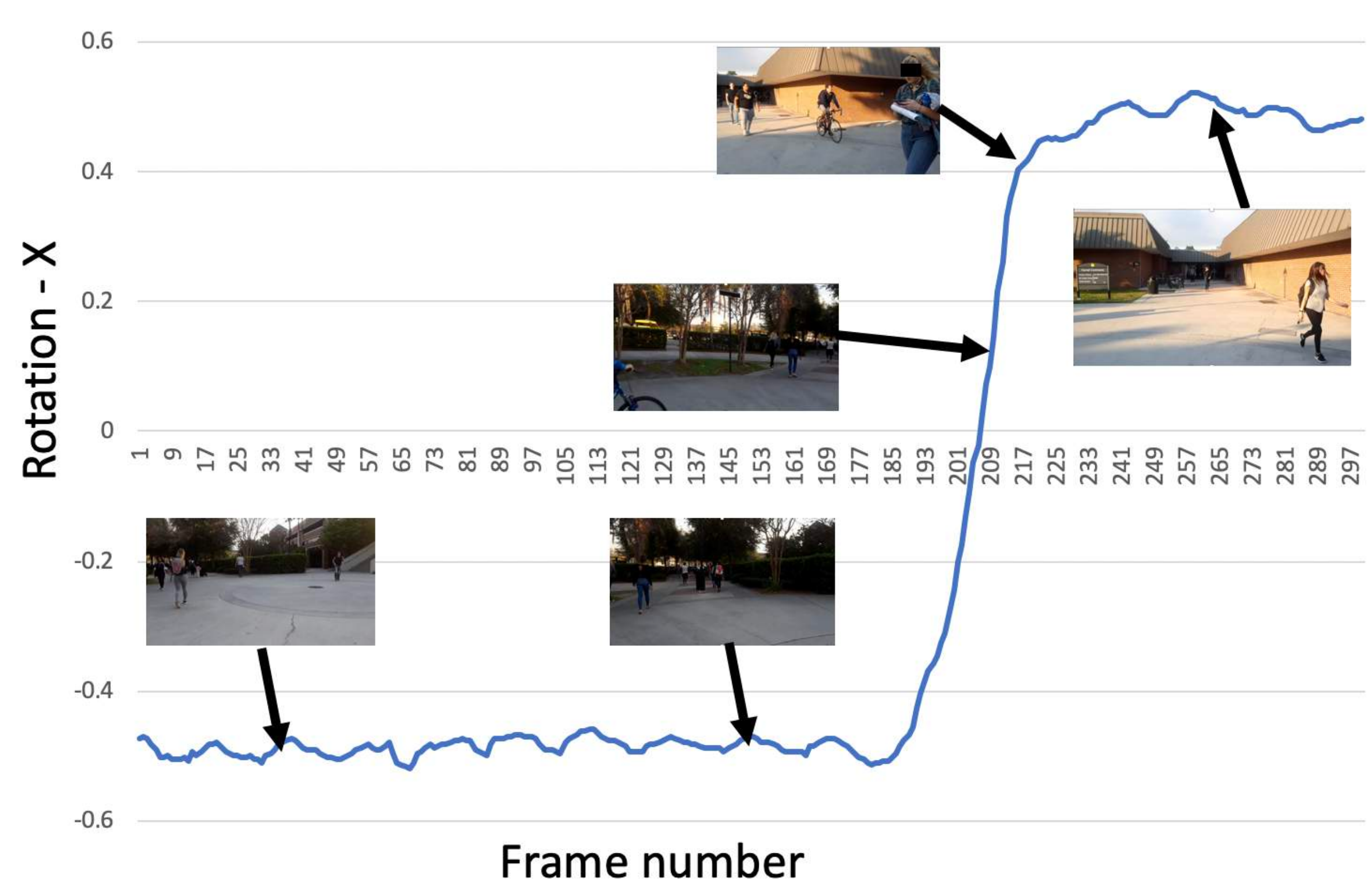} }}%
    \qquad
     \subfloat[]{{\includegraphics[height=4.cm,width=10cm]{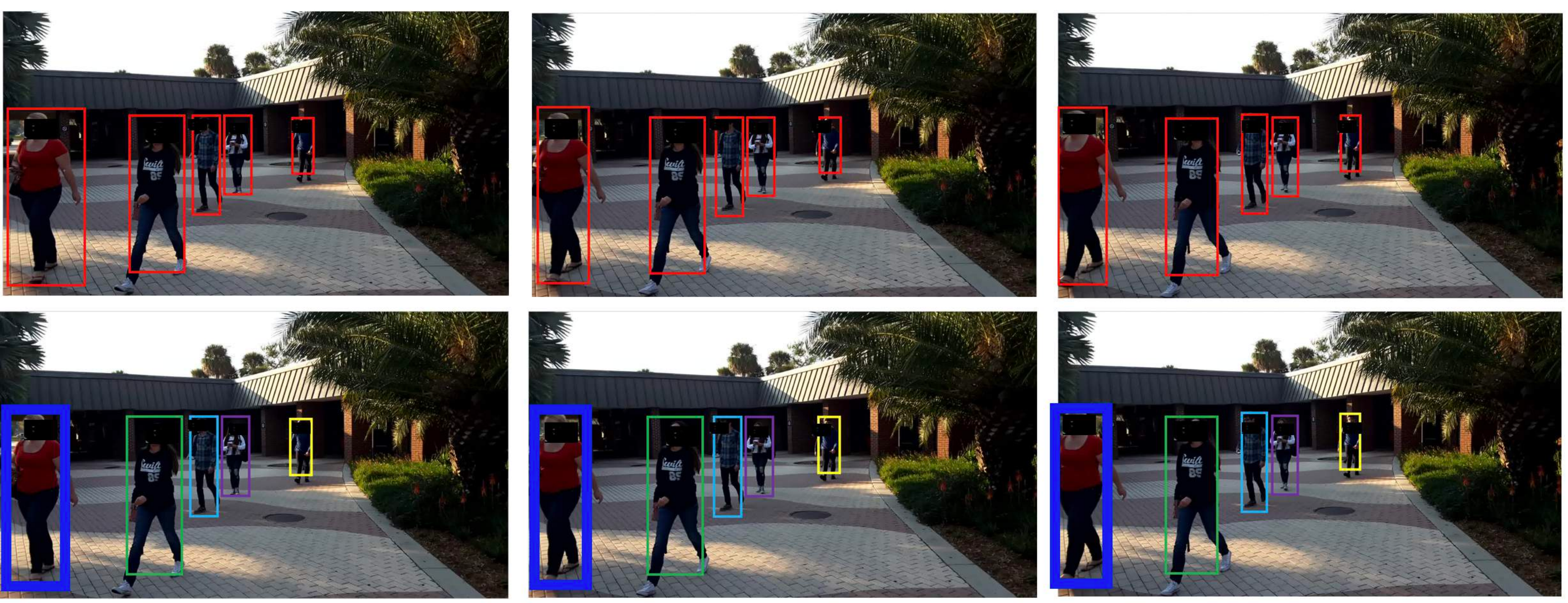}}}%
    \qquad
    \subfloat[]{{\includegraphics[height=4.cm,width=5.2cm]{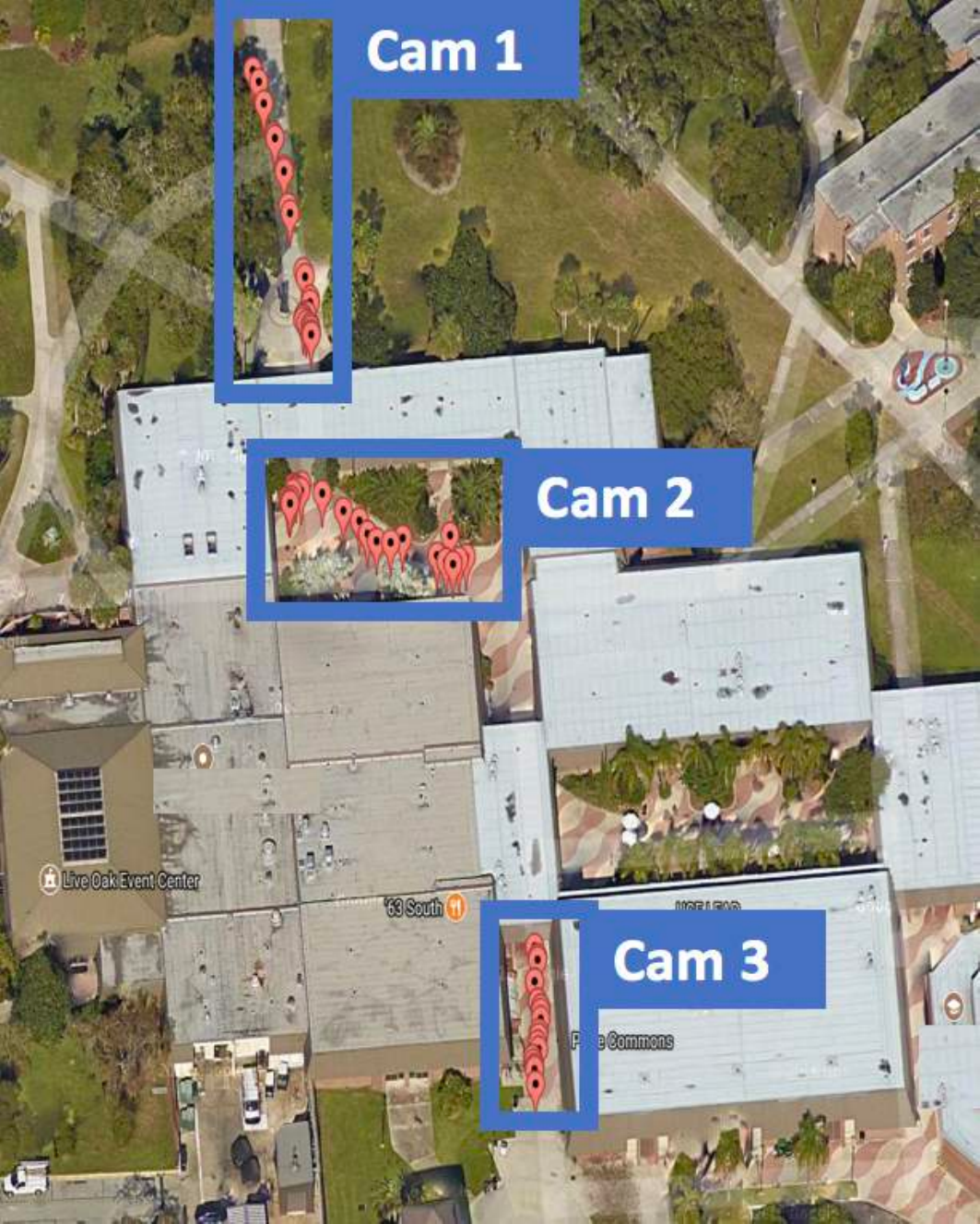}}}%
    \caption{  (a) and (b) respectively show Camera  orientation-pitch Camera rotations around x axis as a functions of time (frame number). We also show corresponding images at different frames. 
    (c) Top row shows sample human detection results and  corresponding tracking results are shown in the bottom row (each track is shown by a different color). (d) shows sample GPS locations of three cameras as  red dots. 
    }
    \label{fig:dataset_plots}%
\end{figure*}

 EgoReID dataset consists of around 900 different identities with 10,200 tracklets with a total of 176,000 human detection bounding boxes. Camera 1 and 2 have 190 IDs in common while camera 2 and 3 share 256 IDs in common. Around 103 IDs are present in all three cameras. 3164, 2748 and 4353 tracklets are present in camera 1,2 and 3, respectively. Each camera has around 36,000 frames. Each pedestrian detection crops have different sizes. The largest crop is 1080 $\times$1496 (width$\times$height), as cameras are very close to pedestrians, they often cover most of the frame. While the smallest detection is 51 $\times$ 38.
 
 Moreover, the dataset contains 12 sensor meta data, namely, Longitude, Latitude, Speed, Distance, Time, Accelerometer, Heading, Gyro, Magnetic, Gravity, Orientation and Rotation vectors. Please refer to Fig. $\ref{fig:dataset_plots}$.
 
 Table \ref{table:Datasets} summarizes the statistics of existing datasets \footnote {Among existing ReID datasets which are collected using fixed camera, we only consider \textbf{video-based} ReID datasets to be included in table \ref{table:Datasets}.} acquired by both fixed and mobile (with First-person POV) cameras. As can be seen from the table, our dataset is the first and large Video-based egocentric dataset. 


\noindent\textbf{Evaluation protocol:} Similar to most of previous datasets, we utilize the Cumulative Matching Characteristics (CMC) curve to evaluate the ReID performance. For each query, multiple true positives could be returned. Therefore, we also consider person ReID as a retrieval task, and also employ  mean Average Precision (mAP) as the evaluation metric.

\section{Proposed Method}
\label{section:method}

\begin{figure*}
    \centering
    \includegraphics[width=\textwidth]{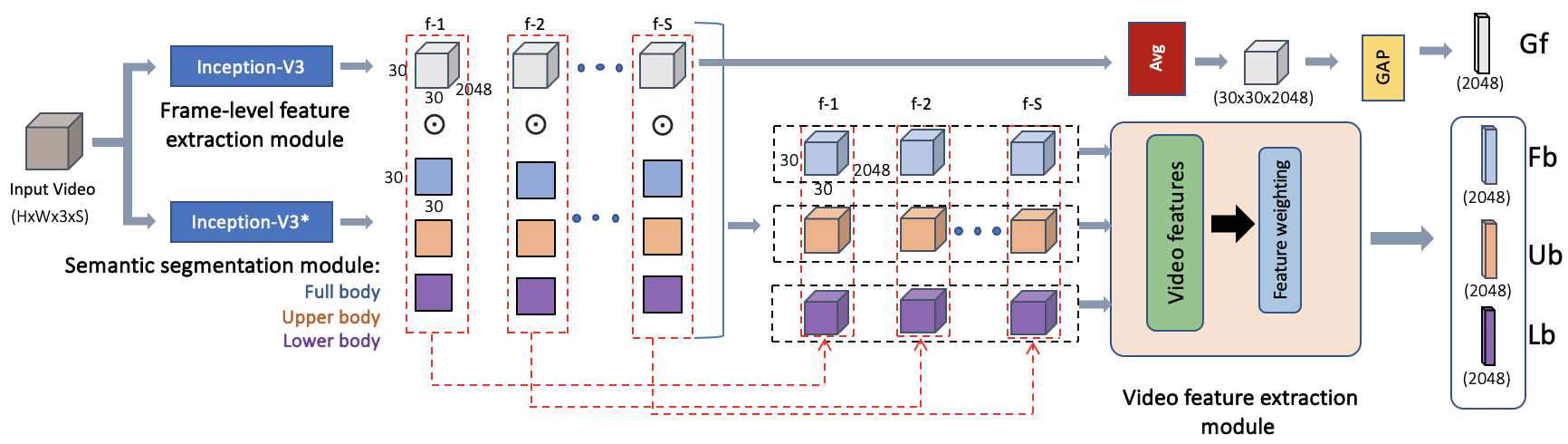}
    \caption{Proposed Network: First,  the input video clip consisting of S frames is input to the Inception-V3 models, the feature maps and 3 segmentation maps for each frame (f-1 to f-S) are generated in the upper and lower branches of the model, respectively. We pool the feature maps using segmentation maps and obtain  sequences of 3 feature maps, each focusing on a different body regions: lower, upper and full. Next, we encode the temporal information within each of these sequences using our video feature extractor and then we scale each feature using their corresponding learned weight. In addition to these feature vectors, we construct another feature representation by taking the average of the output feature maps of Inception-V3 (top branch) and applying global average pooling (GAP) on the resulting tensor. Gf, Fb, Ub and Lb represent global feature, full body, upper body and lower body, respectively 
    }\label{fig:method}
\end{figure*}

This section briefly presents the proposed model which consists of 3 main components: frame-level feature extraction, semantic segmentation and video-level weighted feature extraction. The overview of the proposed network is shown in Figure \ref{fig:method}.

Our model is inspired by the model in \cite{kalayeh2018human}, where human parsing is successfully applied in image-based ReID. Important feature of their model is that it is trained on ten different datasets providing robust system. 
In the unique nature of our dataset, since targets are close to the cameras, different body regions of a target are reasonably  visible, therefore above method is pretty suitable for our problem. However, their method is image-based, in this paper, we extend their method to videos and achieve significantly better results as compared to the existing state-of-the-art methods. 

Since we do not have video level semantic segmentation annotations, it is not possible to 
inflate \cite{carreira2017quo} the bottom 2D layers to 3D, such that network  accepts a video clip instead of an image and generates its  single compact representation. 
we do not have video level semantic segmentation annotations.
Therefore, we apply feature extraction and semantic segmentation at frame level first and then pool each feature map using different segmentation regions.  This is followed by 3D convolution over feature activation map, with attention applied  to each region, to better encode the temporal information in  the video sequence. Finally,   each feature vector is scaled  with its corresponding learned weight. 

Below, first we  describe Frame-level feature Extraction and Semantic Segmentation Modules followed by video feature extraction module.

\subsection{Frame-level feature extraction module:}
In our model, we adopt Inception-V3 \cite{szegedy2016rethinking} as a backbone of our feature  extraction module. Given a video sequence $\mathcal{X} \in \mathbb{R}^{ H\times W \times 3\times S}$ of length $S$, where each frame (RGB) is of size $H \times W$, 
which is passed to Inception-V3 to produce the frame-level feature maps. Feature maps from the Inception-V3,   are then scaled up to the size of the segmentation maps, using bilinear interpolation.

\begin{figure}[]
\begin{center}
  \includegraphics[height=5cm,width=8.0cm]{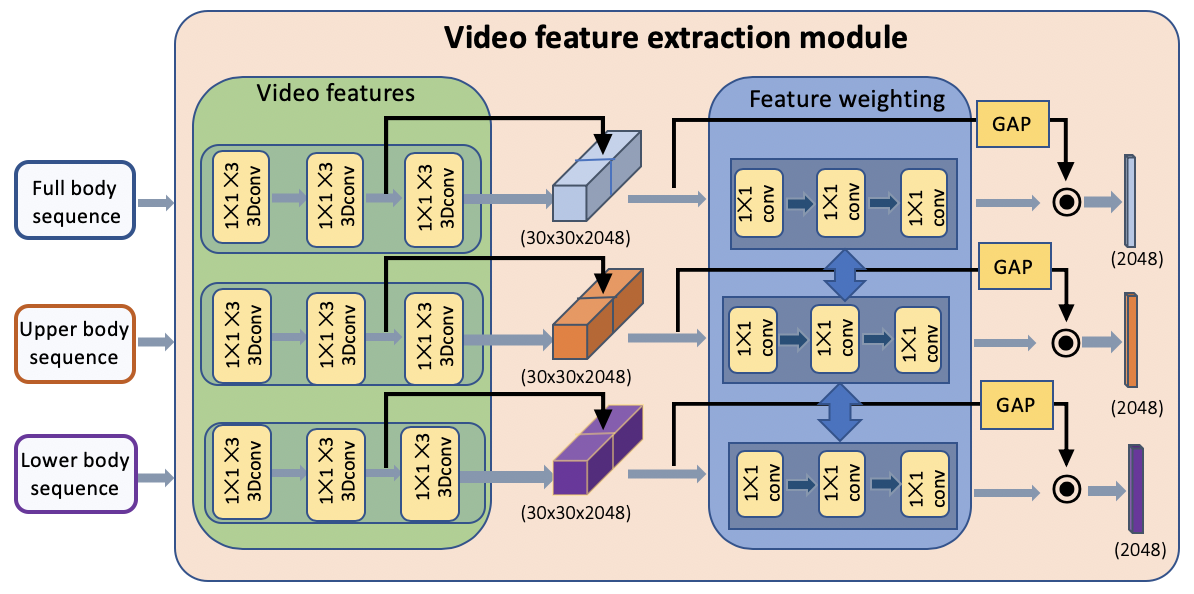}
\end{center}
  \caption{Video feature extraction module: Given three features attentioned to different body regions (upper, lower and full body), we aggregate the temporal information within each of these sequences using separate three $(1\times1\times3)$ 3D-convolution blocks, 
  with spatial and temporal stride of 1 and 2, respectively. 
  We then pass the concatenated features from the last and intermediate layers of video module to our weighting module to learn their respective weights. Finally, after applying global average pooling (GAP) on our final feature maps, we then scale them using their corresponding weights. We share parameters between all three feature weighting blocks.
  }
\label{fig:long}
\label{fig:weight}
\end{figure}


\subsection{Semantic segmentation module:}
 Similar to feature extraction module, for  the segmentation module, we use the Inception-V3 architecture. However, we make two important modification to this architecture  compared to the one used for feature extraction. The first modification is that the output stride of the model is reduced from 32 to 16 to get the feature maps with adequate resolution for semantic segmentation task. 
The extra computation cost resulting from this modification is eliminated by the replacing the convolutions in the last Inception block with the dilated convolutions \cite{yu2015multi}. 
The second modification is the use of atrous spatial pyramid pooling \cite{chen2017rethinking}, instead of global average pooling to exploit multi-scale information. 
Unlike authors in \cite{kalayeh2018human}, where they used 5 body regions and train their ReID model on a single classifier, in our model, we use three body regions: foreground, upper and lower body, and train a separate classifier for each body region. 

\subsection{Video feature extraction module:}
In this module, we first extract video features by aggregating the temporal information within each sequences using a series of 3D-convolution blocks. Then, we learn weights for each region features. Fig. \ref{fig:weight} depicts the proposed video feature extraction module.\\ 

\noindent\textbf{Video features:} An input to this module is three sequences of feature maps each representing different body regions. We then pass them through three separate blocks of 3D-convolutions, each with $1\times1\times3$ kernel size  with spatial and temporal stride of 1 and 2, respectively. 
To exploit multi-scale information, we concatenate feature maps from the intermediate and final layer resulting $(30\times30\times2048)$ feature map representation for each body region. 
We do not share weights between 3D-convolution blocks of different body region.\\


\noindent\textbf{Feature weighting:} In EgoReID, it is more often that different body regions get occluded/absent due to close proximity of camera and target. Thus, it is important that features from different regions are carefully weighted based on their significance. For each video features, we learn their corresponding weights using shared blocks of three 2D-convolution. Finally, after applying global average pooling (GAP) on the feature maps generated by our video feature extractor, we scale them using their corresponding weights. 


During training of the model, we pass the 3 feature vectors of the body regions to different soft-max layers that do not share weights and compute the classification losses separately. 

\section{ Employing Sensor Meta-data}
In this section, we  discuss the proposed method for employing sensor meta data information to further help refine our re-identification results. In particular, we use the heading, speed and GPS (longitude and latitude) of the camera, which are captured by our recording devices. 

\begin{figure}[]
\begin{center}
  \includegraphics[height=6.6cm,width=5.cm]{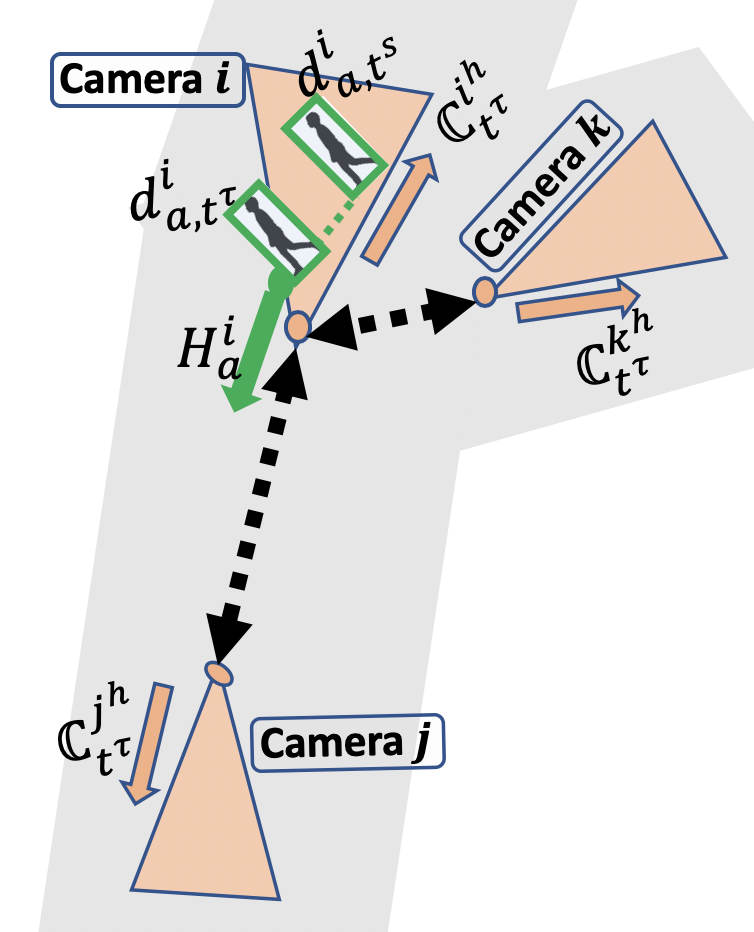}
\end{center}
  \caption{Camera Configuration Example: Lets assume the above topology represents the scene layout at time $t^\tau$ (at the time of exit of a target, shown in green boxes, from camera $i$. The orange triangles represent camera FOVs and their corresponding headings are indicated by the orange arrows. The black doted lines shows the distance between camera $i$ and the other two cameras. The green dotted line and arrow respectively represent a target's tracklet and its heading. In this particular scenario, its evident that camera $j$ should be selected as the \textit{next camera}, since the target is heading towards it, even though camera $k$ is closest camera to the target. And the estimated time is computed using the first case of eq. $\ref{time}$, as both camera $j$ (\textit{next camera}) and target have the same heading. 
  }
\label{fig:ToyExample}
\end{figure}

Let $T^{i}_{a}$  represents tracklet of person $a$  in camera $i$.  
Its trajectory is given by a set of detections $T^{i}_{a}= $ $[d^i_{a,t^{s}}, ..., d^i_{a,t^{\tau}}]$, where $d^i_{a,t^{s}}$ and $d^i_{a,t^{\tau}}$  
represent detections of person $a$ at time of entry and exit from camera $i$, respectively.
$\rho(T^{i}_{a})$ denote appearance feature representation of tracklet $T_a^i$  generated by the proposed model. GPS and heading of camera $i$, $\mathbb{C}^{i}$, at time $t^{\tau}$ are respectively represented by $\mathbb{C}^{i^{g}}_{t^\tau}$ and $\mathbb{C}^{i^{h}}_{t^\tau}$. The heading of a target $a$ in camera $i$, $H^{i}_{a}$, is inferred from the camera $i's$ heading at time $t^\tau$, that is, if target is moving in the same direction as camera (target is walking away from the camera), then $H^{i}_{a} = \mathbb{C}^{i^{h}}_{t^\tau}$. Where as, if a target is moving towards the camera, like the target in Fig. $\ref{fig:ToyExample}$, then $H^{i}_{a} = \mathbb{C}^{i^{h}}_{t^\tau} + 180^0$, that is, we assign the opposite heading w.r.t the camera's heading. The direction of target's movement is determined from the tracklet direction. 

\noindent\textbf{Estimating the next camera:} Na\"ively, one can select the closest camera to the target at time, $t^\tau$, as the next camera without considering the target's direction of motion. As can be seen from the  example in Figure.$\ref{fig:ToyExample}$, such approach will end up selecting the wrong camera $k$, which is located on the other side of the target.
So, first, we need to select candidate cameras which are spatially located in the direction of target's heading. Then, we select the closest camera among those based on their GPS distance from the target at time $t^\tau$. In this set up, we not only select the closest camera but also we ensure the next camera is located along the way of target's motion.
 
Lets define a function $\mathbb{F}(GPS_1, GPS_2)$, which returns heading angle between two $GPS$ points. In particular, it determines  which direction $GPS_2$ is located w.r.t $GPS_1$ and is computed as follows:
     \[\mathbb{F}(GPS_1, GPS_2) = atan2(X,Y),\]
 \noindent where,
 $$X = sin(\lambda_2-\lambda_1)*cos(\varphi_2),$$   $$Y = cos(\varphi_1)*sin(\varphi_2) - sin(\varphi_1)*cos(\varphi_2)*cos(\lambda_2-\lambda_1).$$ \noindent $(\lambda_1, \varphi_1)$ and $(\lambda_2, \varphi_2)$ represent tuple of longitude and latitude of $GPS_1$ and $GPS_2$, respectively.

For a camera, $k$, to be in a candidate set, $\mathcal{S}$, of target $T^i_{a}$, the following should hold: $H^{i}_{a} \equiv \mathbb{F}(\mathbb{C}^{i^{g}}_{t^\tau},\mathbb{C}^{k^{g}}_{t^\tau})$. That is, for a camera $k$ to be selected as a candidate, its heading w.r.t the location of current camera of the target, $\mathbb{C}^{i^{g}}_{t^\tau}$, is the same as target's heading $H^{i}_{a}$. Finally, the next camera for target $a$ in camera $i$ is estimated by selecting the closest camera at time $t^{\tau}$ among the candidate cameras, and is given by; $\argmin_{\mathbb{C}^{j}_{t^\tau}} ~ ||G^{i}_{a} - \mathbb{C}^{j^{g}}_{t^\tau}||, ~ \forall \mathbb{C}^{j}_{t^\tau} \in \mathcal{S},$ 
where, $G^{i}_{a} $ is GPS of target $a$ of camera $i$ at time $t^{\tau}$ and due to very close proximity of the cameras to the targets, we can approximate target's GPS by the camera $i's$ GPS at time $t^{\tau}$, $C^{i^{g}}_{t^\tau}$. 

\noindent\textbf{Estimating time of arrival:} Next, we estimate the time required to travel between camera $i$ (current camera) and $j$ (the next camera). Since our cameras topology is very dynamic (due to camera motion), we implicitly compute the time as a division of distance and speed. The speed of target $a$ in camera $i$, $\mathcal{V}^{i}_{a}$, is estimated to be 1.3 m/s, which is the average walking speed of individual \cite{robin2009specification}. While the speed of camera $j$, $\mathbb{C}^{j^{v}}$, can be inferred from the sensor meta-data. The distance between the next camera, $j$, and the target is computed between their GPS. Formally, the time required by target, $T^{i}_{a}$, to reach camera  $\mathbb{C}^j$, from camera $i$ can be computed as follows: 

\begin{equation}
\label{time}
\mathbb{E}({T^{i}_{a},\mathbb{C}^{j}}) =  ~
\begin{cases}
~  ~   \frac{||G^{i}_{a} - \mathbb{C}_{t^\tau}^{j^{g}}||}{|\mathcal{V}^{i}_{a} - \mathbb{C}^{j^{v}}|}, ~ \mbox{ if } ~ H^{i}_{a} \equiv \mathbb{C}_{t^\tau}^{j^{h}}  \\

\\
~ ~ \frac{||G^{i}_{a} - \mathbb{C}_{t^\tau}^{j^{g}}||}{|\mathcal{V}^{i}_{a} +  \mathbb{C}^{j^{v}}|}, ~  ~ Otherwise
\end{cases}
\end{equation}
\noindent where $H^{i}_{a}$ is the heading of target $a$ at the time of exit, $t^{\tau}$, from camera $i$ and is inferred from its heading of camera $i$ at time $t^\tau$.

Note that in eq. $\ref{time}$, since both target and camera are moving, time computation depends on the direction of their motion. If both are moving in the same direction (i.e. same heading), first case in eq. $\ref{time}$, then we divide the distance by the difference of their speeds, otherwise, we estimate the time by dividing the distance with the sum of their speed. 

Finally, we impose our time constraint on our appearance similarity.  The final affinity between query target, $T^i_{a}$, and the rest of tracks, in camera $\mathbb{C}^{j}$, $T^{j}_{l}, \forall l = 1, ..., |T^j|$, where $|T^j|$ is total number of tracks, is updated as follows: 
\begin{equation}
\label{timeFinal}
M({T^{i}_{a},T^{j}_{l}}) =  ~
\begin{cases}
~  ~  \psi(\rho(T^{i}_{a}), \rho(T^j_{l})) , ~ \mbox{ if } ~ T_{l,t^s}^{j} \geq \mathbb{E}({T^{i}_{a},\mathbb{C}^{j}})\\
\\
~ ~ 0 , ~  ~ Otherwise
\end{cases}
\end{equation}

\noindent where $\psi(\rho(T^{i}_{a}), \rho(T^j_{l}))$ is appearance similarity between two track features, $T_{l,t^s}^{j}$ is an entrance time of track $T^j_l$. 
As can be inferred from eq. $\ref{timeFinal}$, during ReID, we only compare track $T^{i}_{a}$ with tracks in camera $j$ which appears after the estimated time of arrival. 
Thus, we can significantly reduce our search space by pruning several wrong matching which violate our time constraint.


\section{Experiments}

\subsection{Datasets}
We present evaluation on proposed EgoReID dataset and widely used, {\em fixed cameras} dataset, MARS \cite{zheng2016}. MARS consists of 6 cameras and 1261 different pedestrians. There are 625 identities for training and 636 identities for testing.

\textbf{Evaluation settings:} Training/Testing split of EgoReID dataset contains 567 identities for training and 309 identities for testing.  
For MARS dataset, we follow the same Training/Testing split setting as proposed by the authors of the dataset. 
To evaluate performance for each algorithm, we report the Cumulative Matching Characteristic (CMC) metric and mean average precision (mAP). 


\noindent\textbf{Training the Network:}
We first train the semantic segmentation module of our proposed model using Look Into Person (LIP) \cite{gong2017look} dataset. We then freeze the semantic segmentation module for the rest of the training. We train our 
frame-level feature extraction module using the segmentation maps 
produced by semantic segmentation module. Since only 3 semantic regions are used in the model, we group the segmentation maps of different regions to create the segmentation maps for foreground, upper-body and lower-body. During the training of the frame-level feature extraction module, we use a training set consisting of 10 image-based ReID datasets. 
 Next, we fine-tune the frame-level feature extraction module by image-based training using the images from the video datasets (MARS or EgoReID). In the last step of the training, we freeze frame-level feature extraction module and train video feature extraction module using video clips of 15 frames. As mentioned in Section \ref{section:method}, there are three consecutive 3D-convolution layers for each region. After concatenating feature maps from the last and intermediate layer, for each regions we produce a tensor of size $(30\times30\times2048)$. 
We then construct feature vectors from these tensors by applying GAP and then scale them using their corresponding weights. We then compute the loss for each region by performing multi-class classification. Final loss of the model is obtained by taking the average of three losses. More implementation details are provided in the supplementary material. 



\noindent\textbf{Ablation study} We investigate the effect of each component of our model by conducting several experiments. In Table. $\ref{table:ablation}$, we show the results of each component in the proposed network. We evaluate the effects of each body regions and sensor meta data. As can be noted from Table. $\ref{table:ablation}$, lower body features perform worse on EgoReID, while they achieve reasonable result on MARS. This is mainly due to the fact that in EgoReID dataset lower body is frequently missing as the camera is very close to the target. We can also observe from Table. $\ref{table:ablation}$ that, jointly using features from different body regions, leads to improvements in the performance than using only one body region. 

We can also see that the proposed approach for using sensor information to prune several unreliable matches, significantly improves the ReID performance of our approach on EgoReID dataset. As shown in Table. $\ref{table:ablation}$, we are able to improve rank-1 and mAP by around 15\% and 10\%, respectively. 
\begin{table}[h!]
\centering
\resizebox{\columnwidth}{!}{%
\begin{tabular}{c| c c| c |c} 
\hline
 & \multicolumn{2}{|c|}{MARS}& \multicolumn{2}{c}{EgoReID} \\
 \hline
 Methods & R-1 & mAP &  R-1 & mAP \\ [0.5ex] 
 \hline
 \hline
 Upper body (Ub) & 70.76 & 56.61& 26.21 & 22.64 \\
 Lower body (Lb)  & 72.12& 57.91 & 16.02 & 11.78 \\
 Full body (Fb) & 77.22 & 63.71& 28.41 & 25.68 \\
 Ub+Lb+Fb   & 80.01 & 67.96 & 35.16 & 30.51 \\
 Ub+Lb+Fb+Gf   & 83.18 & 72.91& 38.84 & 34.62 \\
 {\bf Ub+Lb+Fb+Gf + Metadata} & -  & -& {\bf 53.02} & {\bf 44.79}  \\[1ex] 
 \hline
\end{tabular}}
\caption{Ablation study of our approach on MARS and EgoReID datasets. Gf are global features from frame-level feature extraction module.}
\label{table:ablation}
\end{table}

We have also evaluated the contributions of our feature weighting module and multi-scale features. As observed from table \ref{table:ablation2}, exploiting multi-scale features from intermediate layers improves our performance by 2\% and 5\% in rank-1 and mAP, respectively. Learning weights for each region features based on their discriminative abilities, improves our result by $\sim$2\% both in rank-1 and mAP. 

To further show the effectiveness of the proposed model, we compare our results with \cite{kalayeh2018human}, where we apply their method on frame by frame bases and then employ average pooling over temporal dimension to generate features from each region. As can be seen from Table. $\ref{table:comp_SPReID}$, our approach gives significantly better results in both rank-1 and mAP (i.e. 24\% and 41\% respectively). This shows the effectiveness of the proposed video feature extractor module.
\begin{table}[h!]
\centering
\resizebox{\columnwidth}{!}{%
\begin{tabular}{c| c c c} 
 \hline
 Methods & R-1 & R-5 & mAP \\ [0.5ex] 
 \hline
 \hline
 SPReID \cite{kalayeh2018human} (Avg pooling) & 58.23 & 72.07 &   31.21 \\ 
 Ours & \textbf{83.18} &\textbf{ 93.28 }& \textbf{72.91} \\[1ex] 
 \hline
\end{tabular}}
\caption{Comparison of SPReID \cite{kalayeh2018human} and the proposed approach on MARS dataset.}
\label{table:comp_SPReID}
\end{table}

\begin{table}[h!]
\centering
\resizebox{\columnwidth}{!}{%
\begin{tabular}{c| c c c} 
 \hline
 Methods & R-1 & R-5 & mAP \\ [0.5ex] 
 \hline
 \hline

Without both weights \& multi-scale features & 79.19 & 89.70 & 65.91 \\[1ex]
With multi-scale features & 81.22 & 92.14 & 70.30 \\[1ex] 
With multi-scale features and weights & \textbf{83.18} & \textbf{93.28} &   \textbf{72.91 }\\ 

 \hline
\end{tabular}}
\caption{Ablation study of different components of our approach on MARS dataset.}
\label{table:ablation2}
\end{table}


\noindent\textbf{Comparison to state-of-the-art Methods:} In Table. $\ref{table:UCFego}$ and Table. $\ref{table:Mars}$, we compare our approach against the state-of-the-art methods on EgoReID and MARS, respectively. %
As can be observed from Table. $\ref{table:UCFego}$, our approach significantly outperforms state-of-the-art approaches on EgoReID dataset. This shows the effectiveness of our human semantic region based local feature extraction approach on our dataset. This is mainly due to  different body parts of pedestrians being clearly visible in EgoReID. In Table. $\ref{table:Mars}$, we observe that our method was able to outperform current stat-of-the-art approaches on MARS dataset in all rank 1,5 and mAP. This further demonstrates the robustness of our approach in handling different domains of video. 
\begin{table}[h!]
\centering
\begin{tabular}{c| c c c } 
 \hline
 Methods & R-1 & R-5 &  mAP  \\ [0.5ex] 
 \hline
 \hline
 PSE+ECN \cite{sarfraz2017pose} & 15.17 & 25.79 & 8.58 \\
 MGCAM \cite{song2018mask}&18.48&29.79&14.60\\
 {\bf Ours} & {\bf 53.02} & {\bf 63.52} &  {\bf 44.79}  \\ [1ex] 
 \hline
\end{tabular}
\caption{Comparison to state-of-the-art on EgoReID dataset. 
}
\label{table:UCFego}
\end{table}


\begin{table}[h!]
\centering
\begin{tabular}{c| c c c } 
 \hline
 Methods & R-1 & R-5 &  mAP  \\ [0.5ex] 
 \hline
 \hline
 
 K-Res \cite{zhong2017re}& 70.51 & - & 55.12 \\ 
 MSCAN \cite{li2017learning}& 71.77 & - & 56.05 \\ 

 SpaAtn \cite{li2018diversity} &  82.30 & - & 65.8 \\ 
 PSE+ ECN \cite{sarfraz2017pose}& 76.70 & - &  71.8  \\
 MGCAM \cite{song2018mask} & 77.17 & - &  71.17  \\
 DuATN \cite{si2018dual} & 81.16 & 92.47 &  67.73  \\
 J. Zhang et al.\cite{zhang2017multi} & 71.20 & 85.70 &  71.8  \\
 \hline
 Ours & 83.18 & 93.28 & 72.91 \\
 Ours+RR & \textbf{85.15} & {\bf 94.95} &  {\bf 81.56}  \\ [1ex] 
 \hline
\end{tabular}
\caption{Comparison to state-of-the-art on MARS dataset. RR is re-ranking using \cite{sarfraz2017pose}.
}
\label{table:Mars}
\end{table}





\section{Conclusion}
In this paper, we presented a new EgoReID dataset which is captured using 3 mobile cellphones with non-overlapping FOV. EgoReID dataset captures substantial variations in lighting,
scene, background, human pose, etc. Compared to the  existing video-based ReID datasets, EgoReID poses several realistic challenges to person ReID task. Unique feature of our dataset is that we also provide 12 sensor meta data for each video. 

We also proposed a new method to solve EgoReID problem, where first frame level local
features are extracted for each semantic region, then 3D
convolutions are applied to encode the temporal information
in each sequence of semantic regions. 
Experiments conducted on MARS and EgoReID showed the effectiveness of our approach in different video domains. In addition, we have also successfully employed sensor meta data information to determine target's next camera and its estimated time of arrival, thus, we only search for a target in the predicted camera around the estimated time of arrival. This significantly improved our ReID performance by reducing our search space.

{\small
\bibliographystyle{ieee}
\bibliography{egbib}

\begin{thebibliography}{10}\itemsep=-1pt

\bibitem{aghaei2016multi}
M.~Aghaei, M.~Dimiccoli, and P.~Radeva.
\newblock Multi-face tracking by extended bag-of-tracklets in egocentric
  photo-streams.
\newblock {\em Computer Vision and Image Understanding}, 149:146--156, 2016.

\bibitem{alahi2008object}
A.~Alahi, M.~Bierlaire, and M.~Kunt.
\newblock Object detection and matching with mobile cameras collaborating with
  fixed cameras.
\newblock In {\em Workshop on Multi-camera and Multi-modal Sensor Fusion
  Algorithms and Applications-M2SFA2 2008}, 2008.

\bibitem{alahi2008master}
A.~Alahi, D.~Marimon, M.~Bierlaire, and M.~Kunt.
\newblock A master-slave approach for object detection and matching with fixed
  and mobile cameras.
\newblock In {\em 15th IEEE International Conference on Image Processing},
  number CONF, 2008.

\bibitem{ardeshir2016ego2top}
S.~Ardeshir and A.~Borji.
\newblock Ego2top: Matching viewers in egocentric and top-view videos.
\newblock In {\em European Conference on Computer Vision}, pages 253--268.
  Springer, 2016.

\bibitem{baltieri20113dpes}
D.~Baltieri, R.~Vezzani, and R.~Cucchiara.
\newblock 3dpes: 3d people dataset for surveillance and forensics.
\newblock In {\em Proceedings of the 2011 joint ACM workshop on Human gesture
  and behavior understanding}, pages 59--64. ACM, 2011.

\bibitem{bengio2013advances}
Y.~Bengio, N.~Boulanger-Lewandowski, and R.~Pascanu.
\newblock Advances in optimizing recurrent networks.
\newblock In {\em 2013 IEEE International Conference on Acoustics, Speech and
  Signal Processing}, pages 8624--8628. IEEE, 2013.

\bibitem{bettadapura2015egocentric}
V.~Bettadapura, I.~Essa, and C.~Pantofaru.
\newblock Egocentric field-of-view localization using first-person
  point-of-view devices.
\newblock In {\em Applications of Computer Vision (WACV), 2015 IEEE Winter
  Conference on}, pages 626--633. IEEE, 2015.

\bibitem{borji2014look}
A.~Borji, D.~N. Sihite, and L.~Itti.
\newblock What/where to look next? modeling top-down visual attention in
  complex interactive environments.
\newblock {\em IEEE Transactions on Systems, Man, and Cybernetics: Systems},
  44(5):523--538, 2014.

\bibitem{carreira2017quo}
J.~Carreira and A.~Zisserman.
\newblock Quo vadis, action recognition? a new model and the kinetics dataset.
\newblock In {\em Computer Vision and Pattern Recognition (CVPR), 2017 IEEE
  Conference on}, pages 4724--4733. IEEE, 2017.

\bibitem{chen2017rethinking}
L.-C. Chen, G.~Papandreou, F.~Schroff, and H.~Adam.
\newblock Rethinking atrous convolution for semantic image segmentation.
\newblock {\em arXiv preprint arXiv:1706.05587}, 2017.

\bibitem{chen2017beyond}
W.~Chen, X.~Chen, J.~Zhang, and K.~Huang.
\newblock Beyond triplet loss: a deep quadruplet network for person
  re-identification.
\newblock In {\em The IEEE Conference on Computer Vision and Pattern
  Recognition (CVPR)}, volume~2, 2017.

\bibitem{dehghan2015gmmcp}
A.~Dehghan, S.~Modiri~Assari, and M.~Shah.
\newblock Gmmcp tracker: Globally optimal generalized maximum multi clique
  problem for multiple object tracking.
\newblock In {\em Proceedings of the IEEE Conference on Computer Vision and
  Pattern Recognition}, pages 4091--4099, 2015.

\bibitem{fathi2011understanding}
A.~Fathi, A.~Farhadi, and J.~M. Rehg.
\newblock Understanding egocentric activities.
\newblock In {\em Computer Vision (ICCV), 2011 IEEE International Conference
  on}, pages 407--414. IEEE, 2011.

\bibitem{fathi2012learning}
A.~Fathi, Y.~Li, and J.~M. Rehg.
\newblock Learning to recognize daily actions using gaze.
\newblock In {\em European Conference on Computer Vision}, pages 314--327.
  Springer, 2012.

\bibitem{fathi2011learning}
A.~Fathi, X.~Ren, and J.~M. Rehg.
\newblock Learning to recognize objects in egocentric activities.
\newblock In {\em Computer Vision and Pattern Recognition (CVPR), 2011 IEEE
  Conference On}, pages 3281--3288. IEEE, 2011.

\bibitem{felzenszwalb2009object}
P.~F. Felzenszwalb, R.~B. Girshick, D.~McAllester, and D.~Ramanan.
\newblock Object detection with discriminatively trained part-based models.
\newblock {\em IEEE transactions on pattern analysis and machine intelligence},
  32(9):1627--1645, 2009.

\bibitem{fergnani2016body}
F.~Fergnani, S.~Alletto, G.~Serra, J.~De~Mira, and R.~Cucchiara.
\newblock Body part based re-identification from an egocentric perspective.
\newblock In {\em Proceedings of the IEEE Conference on Computer Vision and
  Pattern Recognition Workshops}, pages 1--6, 2016.

\bibitem{ferland2009egocentric}
F.~Ferland, F.~Pomerleau, C.~T. Le~Dinh, and F.~Michaud.
\newblock Egocentric and exocentric teleoperation interface using real-time, 3d
  video projection.
\newblock In {\em Proceedings of the 4th ACM/IEEE international conference on
  Human robot interaction}, pages 37--44. ACM, 2009.

\bibitem{gong2017look}
K.~Gong, X.~Liang, D.~Zhang, X.~Shen, and L.~Lin.
\newblock Look into person: Self-supervised structure-sensitive learning and a
  new benchmark for human parsing.
\newblock In {\em CVPR}, volume~2, page~6, 2017.

\bibitem{gong2014person}
S.~Gong, M.~Cristani, S.~Yan, and C.~C. Loy.
\newblock {\em Person re-identification}.
\newblock Springer, 2014.

\bibitem{gray2007evaluating}
D.~Gray, S.~Brennan, and H.~Tao.
\newblock Evaluating appearance models for recognition, reacquisition, and
  tracking.
\newblock In {\em Proc. IEEE International Workshop on Performance Evaluation
  for Tracking and Surveillance (PETS)}, volume~3, pages 1--7. Citeseer, 2007.

\bibitem{hirzer2011person}
M.~Hirzer, C.~Beleznai, P.~M. Roth, and H.~Bischof.
\newblock Person re-identification by descriptive and discriminative
  classification.
\newblock In {\em Scandinavian conference on Image analysis}, pages 91--102.
  Springer, 2011.

\bibitem{jiang2017seeing}
H.~Jiang and K.~Grauman.
\newblock Seeing invisible poses: Estimating 3d body pose from egocentric
  video.
\newblock In {\em 2017 IEEE Conference on Computer Vision and Pattern
  Recognition (CVPR)}, pages 3501--3509. IEEE, 2017.

\bibitem{kalayeh2018human}
M.~M. Kalayeh, E.~Basaran, M.~G{\"o}kmen, M.~E. Kamasak, and M.~Shah.
\newblock Human semantic parsing for person re-identification.
\newblock In {\em Proceedings of the IEEE Conference on Computer Vision and
  Pattern Recognition}, pages 1062--1071, 2018.

\bibitem{kawanishi2014shinpuhkan2014}
Y.~Kawanishi, Y.~Wu, M.~Mukunoki, and M.~Minoh.
\newblock Shinpuhkan2014: A multi-camera pedestrian dataset for tracking people
  across multiple cameras.
\newblock In {\em 20th Korea-Japan Joint Workshop on Frontiers of Computer
  Vision}, volume~5. Citeseer, 2014.

\bibitem{li2017learning}
D.~Li, X.~Chen, Z.~Zhang, and K.~Huang.
\newblock Learning deep context-aware features over body and latent parts for
  person re-identification.
\newblock In {\em Proceedings of the IEEE Conference on Computer Vision and
  Pattern Recognition}, pages 384--393, 2017.

\bibitem{li2018diversity}
S.~Li, S.~Bak, P.~Carr, and X.~Wang.
\newblock Diversity regularized spatiotemporal attention for video-based person
  re-identification.
\newblock In {\em Proceedings of the IEEE Conference on Computer Vision and
  Pattern Recognition}, pages 369--378, 2018.

\bibitem{li2017person}
S.~Li, T.~Xiao, H.~Li, B.~Zhou, D.~Yue, and X.~Wang.
\newblock Person search with natural language description.
\newblock {\em arXiv preprint arXiv:1702.05729}, 2017.

\bibitem{li2013locally}
W.~Li and X.~Wang.
\newblock Locally aligned feature transforms across views.
\newblock In {\em Proceedings of the IEEE Conference on Computer Vision and
  Pattern Recognition}, pages 3594--3601, 2013.

\bibitem{li2012human}
W.~Li, R.~Zhao, and X.~Wang.
\newblock Human reidentification with transferred metric learning.
\newblock In {\em Asian Conference on Computer Vision}, pages 31--44. Springer,
  2012.

\bibitem{li2014deepreid}
W.~Li, R.~Zhao, T.~Xiao, and X.~Wang.
\newblock Deepreid: Deep filter pairing neural network for person
  re-identification.
\newblock In {\em Proceedings of the IEEE Conference on Computer Vision and
  Pattern Recognition}, pages 152--159, 2014.

\bibitem{li2013learning}
Y.~Li, A.~Fathi, and J.~M. Rehg.
\newblock Learning to predict gaze in egocentric video.
\newblock In {\em 2013 IEEE International Conference on Computer Vision}, pages
  3216--3223. IEEE, 2013.

\bibitem{liu2016spatio}
J.~Liu, A.~Shahroudy, D.~Xu, and G.~Wang.
\newblock Spatio-temporal lstm with trust gates for 3d human action
  recognition.
\newblock In {\em European Conference on Computer Vision}, pages 816--833.
  Springer, 2016.

\bibitem{loy2009}
C.~C. Loy, T.~Xiang, and S.~Gong.
\newblock Multi-camera activity correlation analysis.
\newblock 2009.

\bibitem{lu2013story}
Z.~Lu and K.~Grauman.
\newblock Story-driven summarization for egocentric video.
\newblock In {\em 2013 IEEE Conference on Computer Vision and Pattern
  Recognition}, pages 2714--2721. IEEE, 2013.

\bibitem{ma2017person}
X.~Ma, X.~Zhu, S.~Gong, X.~Xie, J.~Hu, K.-M. Lam, and Y.~Zhong.
\newblock Person re-identification by unsupervised video matching.
\newblock {\em Pattern Recognition}, 65:197--210, 2017.

\bibitem{mclaughlin2016recurrent}
N.~McLaughlin, J.~Martinez~del Rincon, and P.~Miller.
\newblock Recurrent convolutional network for video-based person
  re-identification.
\newblock In {\em Proceedings of the IEEE conference on computer vision and
  pattern recognition}, pages 1325--1334, 2016.

\bibitem{redmon2017yolo9000}
J.~Redmon and A.~Farhadi.
\newblock Yolo9000: better, faster, stronger.
\newblock {\em arXiv preprint}, 2017.

\bibitem{ristani2016performance}
E.~Ristani, F.~Solera, R.~Zou, R.~Cucchiara, and C.~Tomasi.
\newblock Performance measures and a data set for multi-target, multi-camera
  tracking.
\newblock In {\em European Conference on Computer Vision}, pages 17--35.
  Springer, 2016.

\bibitem{robin2009specification}
T.~Robin, G.~Antonini, M.~Bierlaire, and J.~Cruz.
\newblock Specification, estimation and validation of a pedestrian walking
  behavior model.
\newblock {\em Transportation Research Part B: Methodological}, 43(1):36--56,
  2009.

\bibitem{russakovsky2015imagenet}
O.~Russakovsky, J.~Deng, H.~Su, J.~Krause, S.~Satheesh, S.~Ma, Z.~Huang,
  A.~Karpathy, A.~Khosla, M.~Bernstein, et~al.
\newblock Imagenet large scale visual recognition challenge.
\newblock {\em International Journal of Computer Vision}, 115(3):211--252,
  2015.

\bibitem{sarfraz2017pose}
M.~Saquib~Sarfraz, A.~Schumann, A.~Eberle, and R.~Stiefelhagen.
\newblock A pose-sensitive embedding for person re-identification with expanded
  cross neighborhood re-ranking.
\newblock In {\em Proceedings of the IEEE Conference on Computer Vision and
  Pattern Recognition}, pages 420--429, 2018.

\bibitem{si2018dual}
J.~Si, H.~Zhang, C.-G. Li, J.~Kuen, X.~Kong, A.~C. Kot, and G.~Wang.
\newblock Dual attention matching network for context-aware feature sequence
  based person re-identification.
\newblock {\em arXiv preprint arXiv:1803.09937}, 2018.

\bibitem{song2018mask}
C.~Song, Y.~Huang, W.~Ouyang, and L.~Wang.
\newblock Mask-guided contrastive attention model for person re-identification.
\newblock In {\em Proceedings of the IEEE Conference on Computer Vision and
  Pattern Recognition}, pages 1179--1188, 2018.

\bibitem{szegedy2016rethinking}
C.~Szegedy, V.~Vanhoucke, S.~Ioffe, J.~Shlens, and Z.~Wojna.
\newblock Rethinking the inception architecture for computer vision.
\newblock In {\em Proceedings of the IEEE Conference on Computer Vision and
  Pattern Recognition}, pages 2818--2826, 2016.

\bibitem{Tesfaye2019}
Y.~T. Tesfaye, E.~Zemene, A.~Prati, M.~Pelillo, and M.~Shah.
\newblock Multi-target tracking in multiple non-overlapping cameras using
  fast-constrained dominant sets.
\newblock {\em International Journal of Computer Vision}, May 2019.

\bibitem{varior2016gated}
R.~R. Varior, M.~Haloi, and G.~Wang.
\newblock Gated siamese convolutional neural network architecture for human
  re-identification.
\newblock In {\em European Conference on Computer Vision}, pages 791--808.
  Springer, 2016.

\bibitem{wang2016joint}
F.~Wang, W.~Zuo, L.~Lin, D.~Zhang, and L.~Zhang.
\newblock Joint learning of single-image and cross-image representations for
  person re-identification.
\newblock In {\em Proceedings of the IEEE Conference on Computer Vision and
  Pattern Recognition}, pages 1288--1296, 2016.

\bibitem{wang2016person}
T.~Wang, S.~Gong, X.~Zhu, and S.~Wang.
\newblock Person re-identification by discriminative selection in video
  ranking.
\newblock {\em IEEE Trans. Pattern Anal. Mach. Intell.}, 38(12):2501--2514,
  2016.

\bibitem{wang2013}
X.~Wang.
\newblock Intelligent multi-camera video surveillance: A review.
\newblock {\em Pattern recognition letters}, 34(1):3--19, 2013.

\bibitem{xiao2016end}
T.~Xiao, S.~Li, B.~Wang, L.~Lin, and X.~Wang.
\newblock End-to-end deep learning for person search.
\newblock {\em arXiv preprint arXiv:1604.01850}, 2, 2016.

\bibitem{xu2017jointly}
S.~Xu, Y.~Cheng, K.~Gu, Y.~Yang, S.~Chang, and P.~Zhou.
\newblock Jointly attentive spatial-temporal pooling networks for video-based
  person re-identification.
\newblock {\em arXiv preprint arXiv:1708.02286}, 2017.

\bibitem{yao2019unsupervised}
Y.~Yao, M.~Xu, Y.~Wang, D.~J. Crandall, and E.~M. Atkins.
\newblock Unsupervised traffic accident detection in first-person videos.
\newblock {\em arXiv preprint arXiv:1903.00618}, 2019.

\bibitem{you2016top}
J.~You, A.~Wu, X.~Li, and W.-S. Zheng.
\newblock Top-push video-based person re-identification.
\newblock In {\em Proceedings of the IEEE Conference on Computer Vision and
  Pattern Recognition}, pages 1345--1353, 2016.

\bibitem{yu2015multi}
F.~Yu and V.~Koltun.
\newblock Multi-scale context aggregation by dilated convolutions.
\newblock {\em arXiv preprint arXiv:1511.07122}, 2015.

\bibitem{yu2013harry}
S.-I. Yu, Y.~Yang, and A.~Hauptmann.
\newblock Harry potter's marauder's map: Localizing and tracking multiple
  persons-of-interest by nonnegative discretization.
\newblock In {\em 2013 IEEE Conference on Computer Vision and Pattern
  Recognition}, pages 3714--3720. IEEE, 2013.

\bibitem{zhang2017multi}
J.~Zhang, N.~Wang, and L.~Zhang.
\newblock Multi-shot pedestrian re-identification via sequential decision
  making.
\newblock {\em arXiv preprint arXiv:1712.07257}, 2017.

\bibitem{zhao2017spindle}
H.~Zhao, M.~Tian, S.~Sun, J.~Shao, J.~Yan, S.~Yi, X.~Wang, and X.~Tang.
\newblock Spindle net: Person re-identification with human body region guided
  feature decomposition and fusion.
\newblock In {\em Proceedings of the IEEE Conference on Computer Vision and
  Pattern Recognition}, pages 1077--1085, 2017.

\bibitem{zheng2016}
L.~Zheng, Z.~Bie, Y.~Sun, J.~Wang, C.~Su, S.~Wang, and Q.~Tian.
\newblock Mars: A video benchmark for large-scale person re-identification.
\newblock In {\em European Conference on Computer Vision}, pages 868--884.
  Springer, 2016.

\bibitem{zheng2016mars}
L.~Zheng, Z.~Bie, Y.~Sun, J.~Wang, C.~Su, S.~Wang, and Q.~Tian.
\newblock Mars: A video benchmark for large-scale person re-identification.
\newblock In {\em European Conference on Computer Vision}, pages 868--884.
  Springer, 2016.

\bibitem{zheng2015scalable}
L.~Zheng, L.~Shen, L.~Tian, S.~Wang, J.~Wang, and Q.~Tian.
\newblock Scalable person re-identification: A benchmark.
\newblock In {\em Proceedings of the IEEE International Conference on Computer
  Vision}, pages 1116--1124, 2015.

\bibitem{zheng2016person}
L.~Zheng, Y.~Yang, and A.~G. Hauptmann.
\newblock Person re-identification: Past, present and future.
\newblock {\em arXiv preprint arXiv:1610.02984}, 2016.

\bibitem{zheng2017person}
L.~Zheng, H.~Zhang, S.~Sun, M.~Chandraker, Y.~Yang, Q.~Tian, et~al.
\newblock Person re-identification in the wild.
\newblock In {\em CVPR}, volume~1, page~2, 2017.

\bibitem{zheng2017unlabeled}
Z.~Zheng, L.~Zheng, and Y.~Yang.
\newblock Unlabeled samples generated by gan improve the person
  re-identification baseline in vitro.
\newblock {\em arXiv preprint arXiv:1701.07717}, 3, 2017.

\bibitem{zhong2017re}
Z.~Zhong, L.~Zheng, D.~Cao, and S.~Li.
\newblock Re-ranking person re-identification with k-reciprocal encoding.
\newblock In {\em Computer Vision and Pattern Recognition (CVPR), 2017 IEEE
  Conference on}, pages 3652--3661. IEEE, 2017.

\bibitem{zhou2017see}
Z.~Zhou, Y.~Huang, W.~Wang, L.~Wang, and T.~Tan.
\newblock See the forest for the trees: Joint spatial and temporal recurrent
  neural networks for video-based person re-identification.
\newblock In {\em Computer Vision and Pattern Recognition (CVPR), 2017 IEEE
  Conference on}, pages 6776--6785. IEEE, 2017.

\bibitem{zhu2018video}
X.~Zhu, X.-Y. Jing, X.~You, X.~Zhang, and T.~Zhang.
\newblock Video-based person re-identification by simultaneously learning
  intra-video and inter-video distance metrics.
\newblock {\em IEEE Transactions on Image Processing}, 27(11):5683--5695, 2018.

\end{thebibliography}


\begin{thebibliography}{1}\itemsep=-1pt

\bibitem{Alpher02}
A.~Alpher.
\newblock Frobnication.
\newblock {\em Journal of Foo}, 12(1):234--778, 2002.

\bibitem{Alpher03}
A.~Alpher and J.~P.~N. Fotheringham-Smythe.
\newblock Frobnication revisited.
\newblock {\em Journal of Foo}, 13(1):234--778, 2003.

\bibitem{Alpher04}
A.~Alpher, J.~P.~N. Fotheringham-Smythe, and G.~Gamow.
\newblock Can a machine frobnicate?
\newblock {\em Journal of Foo}, 14(1):234--778, 2004.

\bibitem{Authors06b}
Authors.
\newblock Frobnication tutorial, 2006.
\newblock Supplied as additional material {\tt tr.pdf}.

\bibitem{Authors06}
Authors.
\newblock The frobnicatable foo filter, 2011.
\newblock Face and Gesture submission ID 324. Supplied as additional material
  {\tt fg324.pdf}.

\end{thebibliography}
}

\appendix 
\section{Appendices}

We supplement our main submission in four aspects. \textbf{First}, we provide several details on our dataset. We have included plots, sample tracklets and videos to help better explain our EgoReID dataset. \textbf{Second}, to demonstrate the quality of our human semantic parsing model, we have included videos which show qualitative results of human semantic segmentation of sample tracklets. \textbf{Third}, to further elaborate the effectiveness of the proposed person ReID model, we have included qualitative results of some randomly selected queries. \textbf{Forth}, several implementation details are included.

\section{EgoReID Dataset}
\begin{figure*}[t]
\begin{center}
\includegraphics[width=\textwidth]{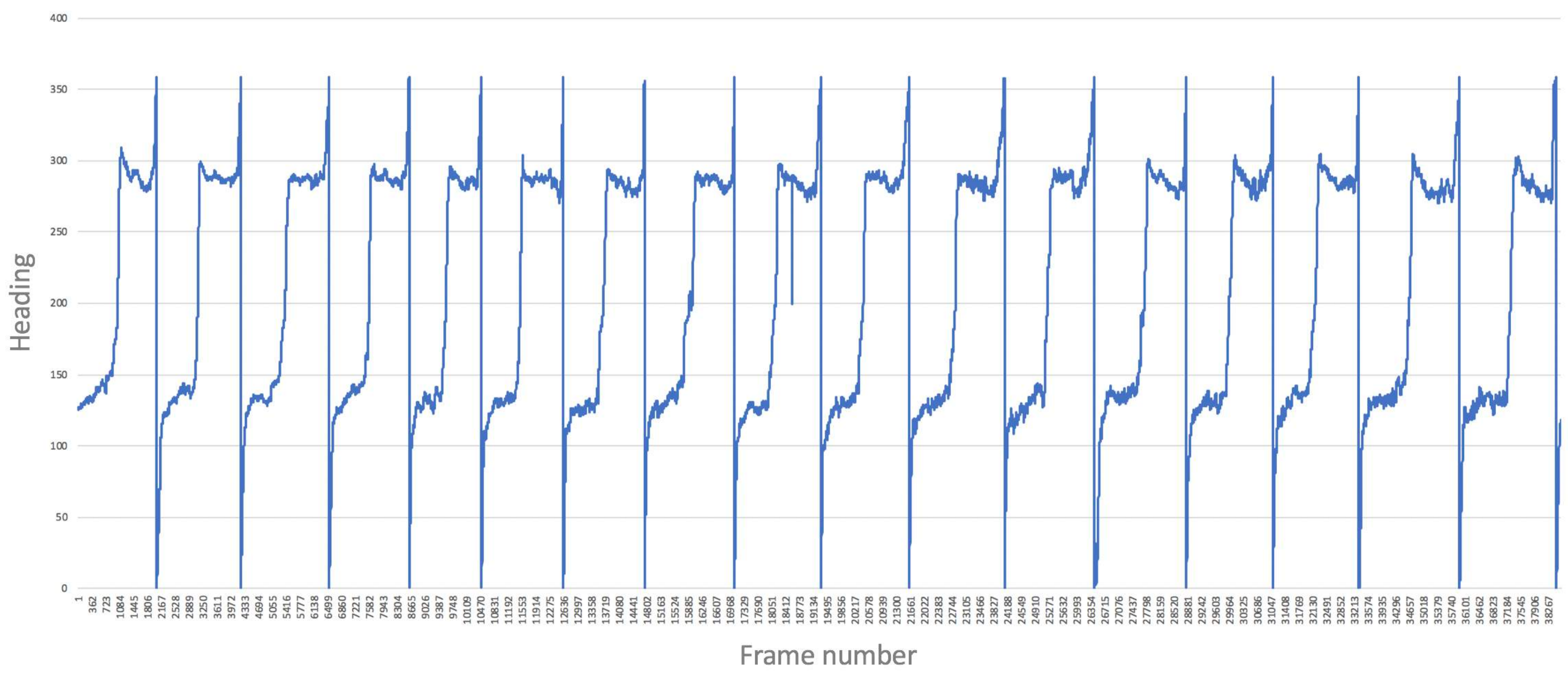}
\end{center}
  \caption{Heading plot.}
\label{fig:heading}
\end{figure*}

\begin{figure*}[t]
\begin{center}
\includegraphics[width=\textwidth]{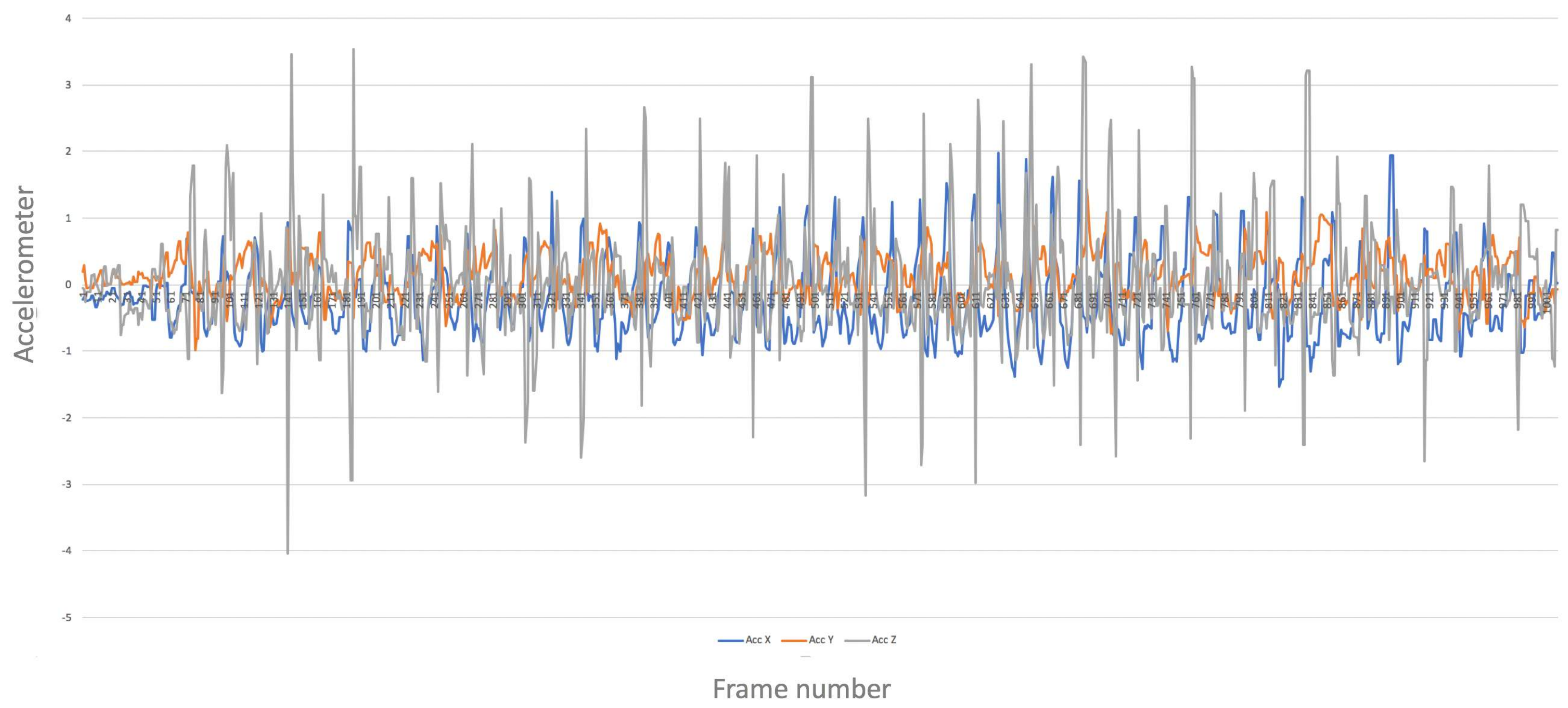}
\end{center}
  \caption{ Accelerometer (x,y,z) plot.}
\label{fig:acc}
\end{figure*}

\begin{figure*}[t]
\begin{center}
\includegraphics[width=\textwidth]{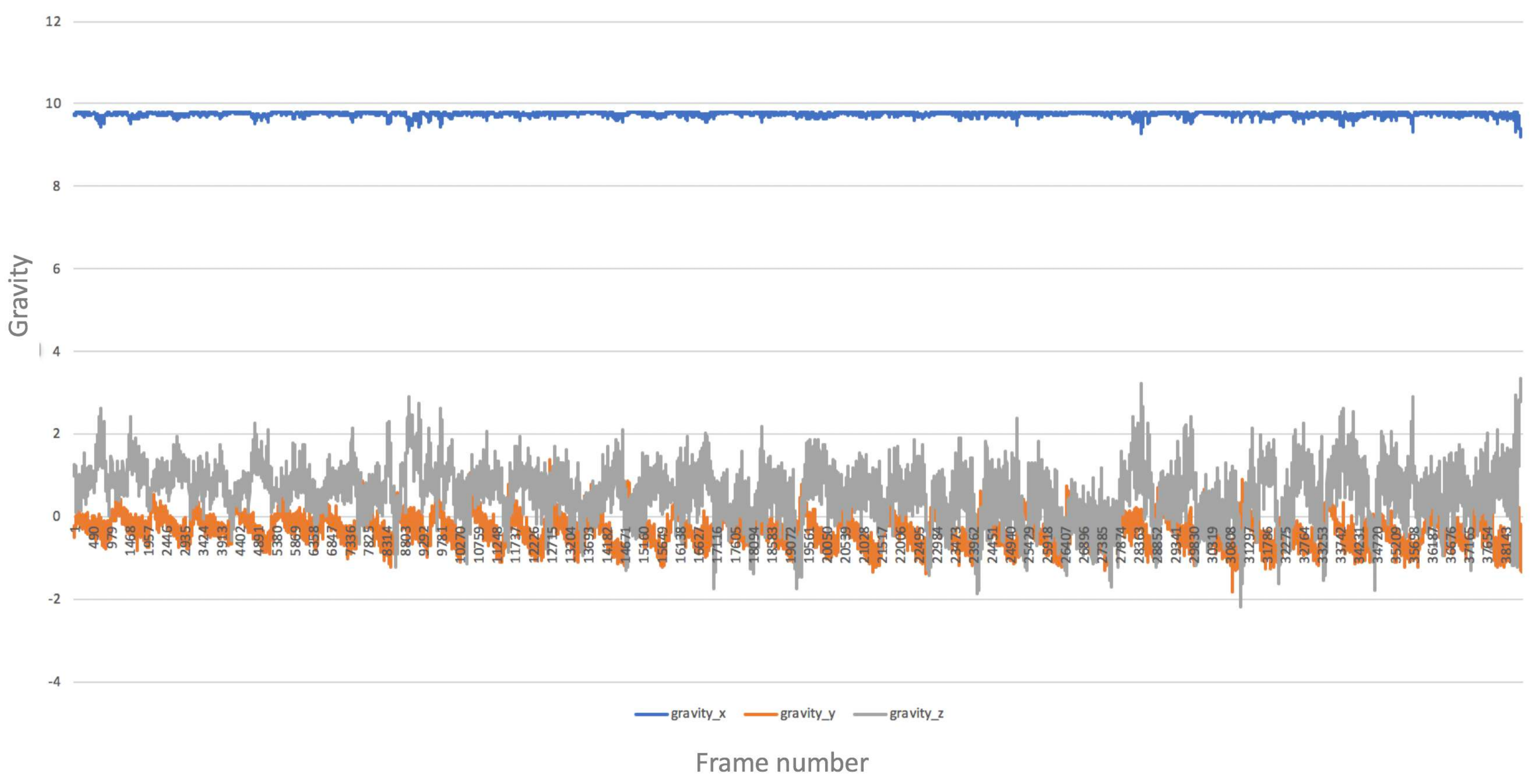}
\end{center}
  \caption{Gravity (x,y,z) Plot.}
\label{fig:gravity}
\end{figure*}

\begin{figure*}[t]
\begin{center}
\includegraphics[width=\textwidth]{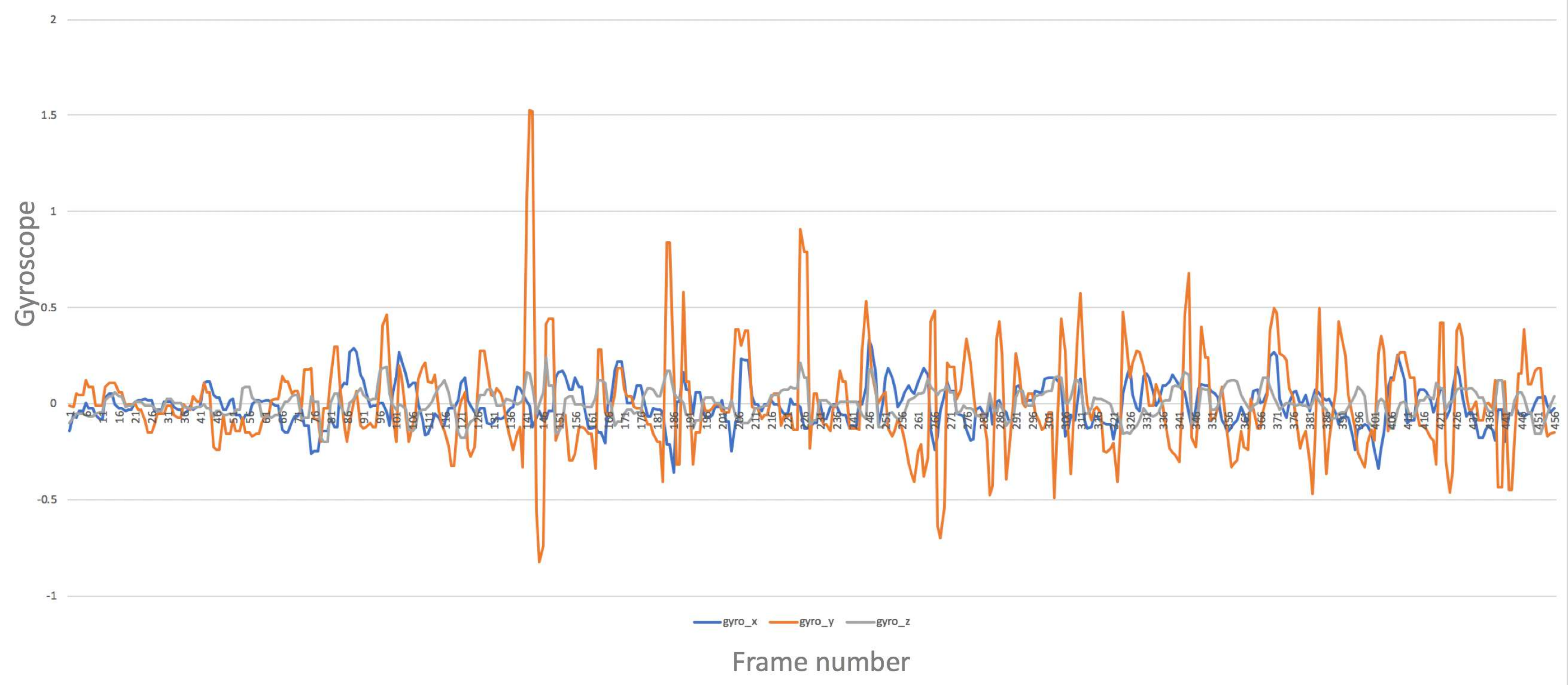}
\end{center}
  \caption{Gyroscope (x,y,z) plot.}
\label{fig:gyro}
\end{figure*}

\begin{figure*}[t]
\begin{center}
\includegraphics[width=\textwidth]{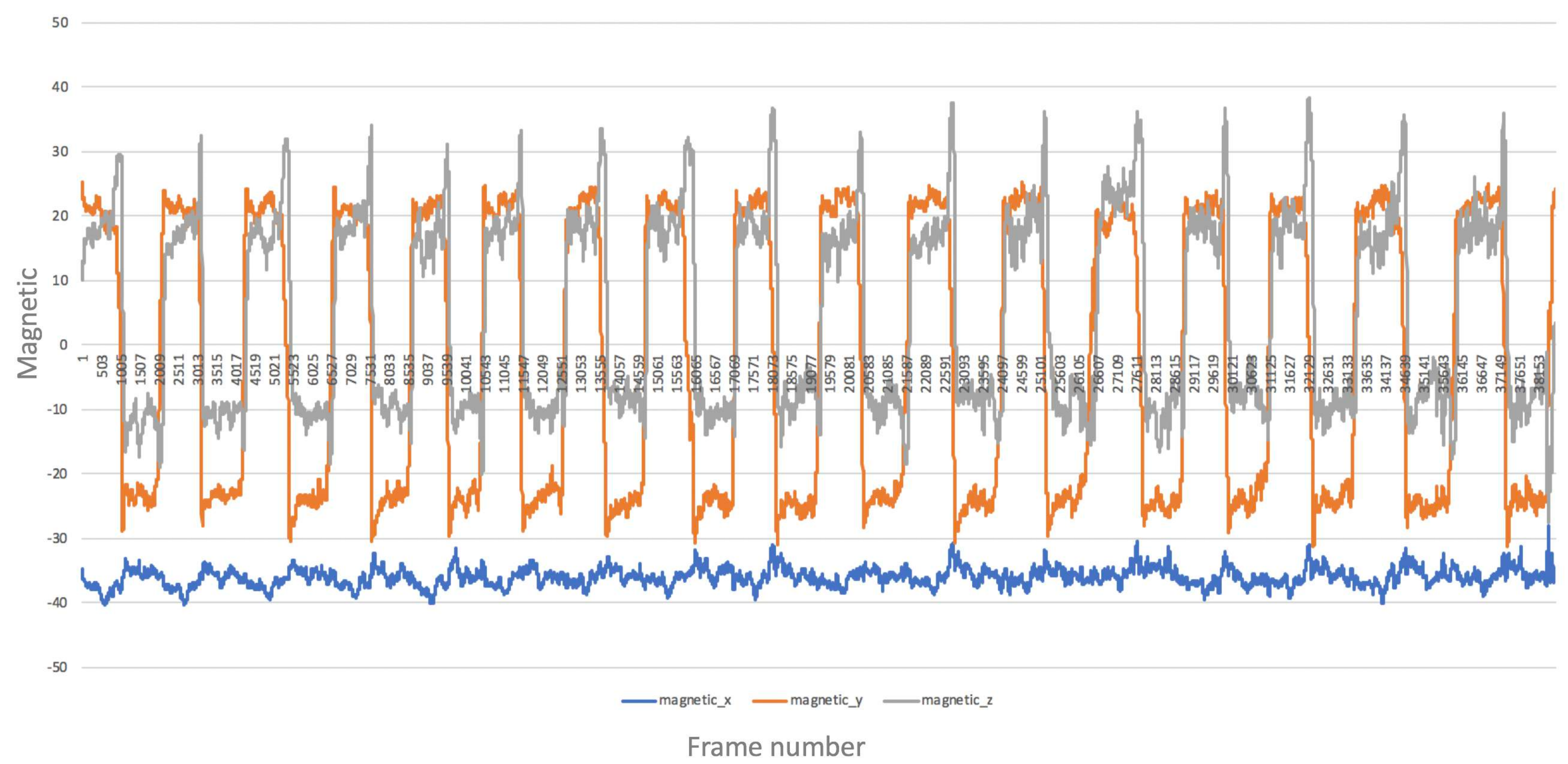}
\end{center}
  \caption{Magnetic (x,y,z) Plot.}
\label{fig:magnetic}
\end{figure*}

\begin{figure*}[t]
\begin{center}
\includegraphics[width=\textwidth]{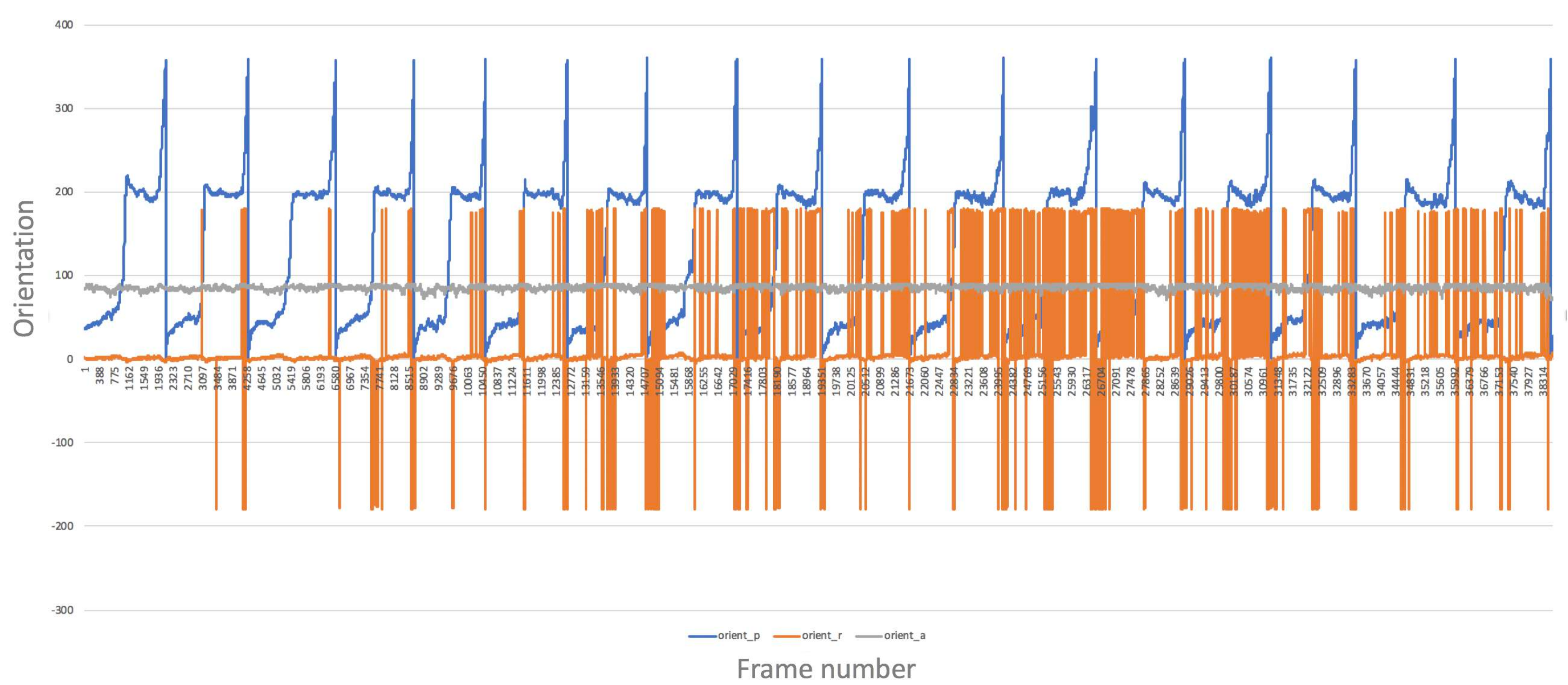}
\end{center}
  \caption{Orientation (x,y,z) plot.}
\label{fig:orientation}
\end{figure*}

\begin{figure*}[t]
\begin{center}
\includegraphics[width=\textwidth]{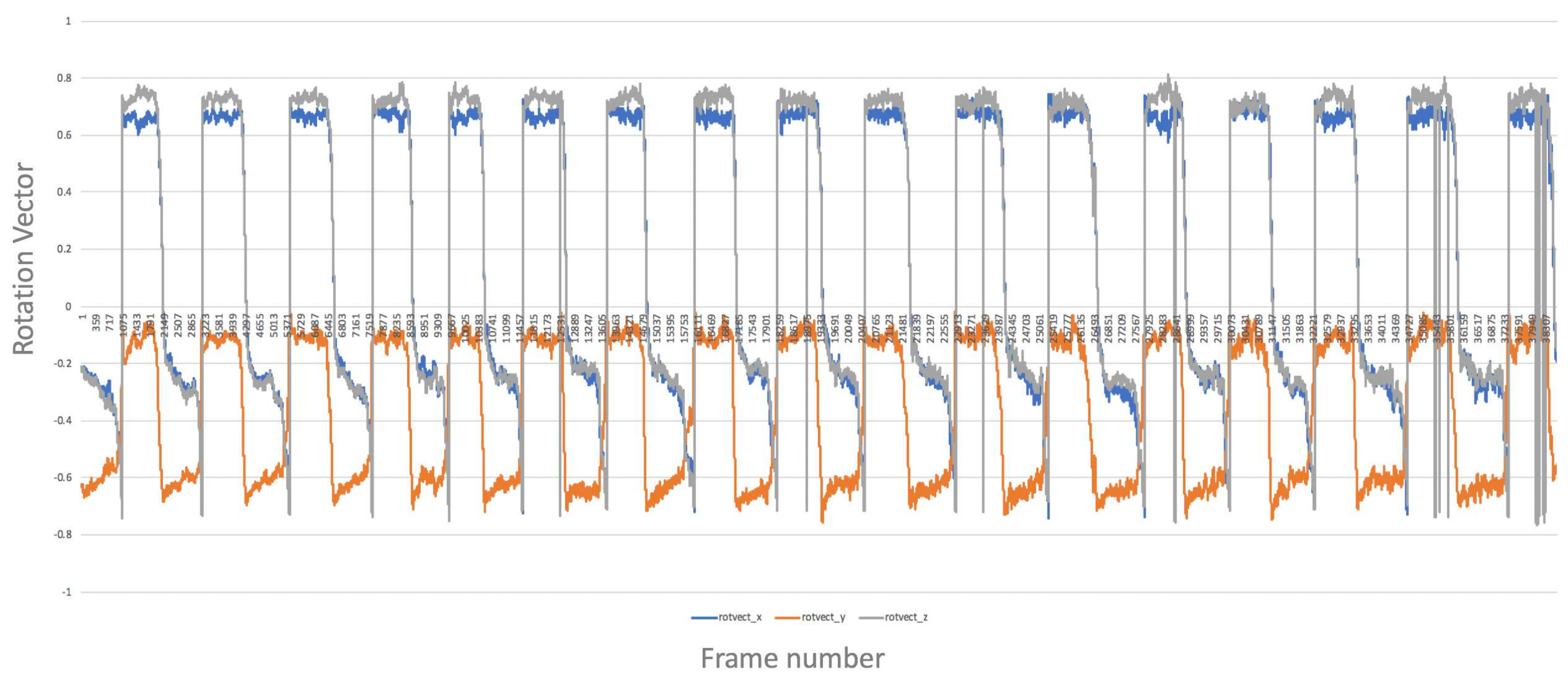}
\end{center}
  \caption{Rotation\_Vector (x,y,z) plot.}
\label{fig:rotation}
\end{figure*}

\begin{figure*}[t]
\begin{center}
\includegraphics[width=\textwidth]{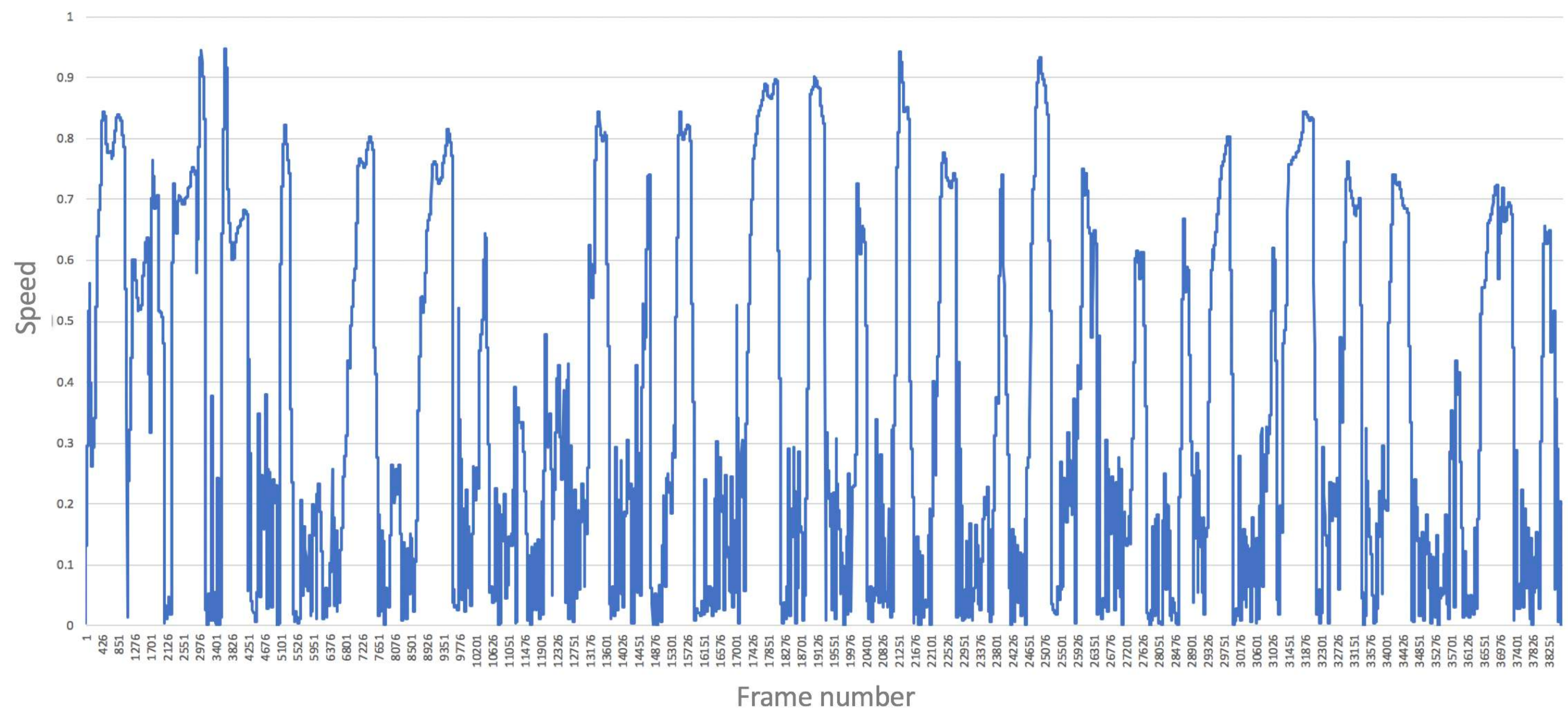}
\end{center}
  \caption{Speed plot.}
\label{fig:speed}
\end{figure*}

In fig. $\ref{fig:heading}$ - $\ref{fig:speed}$, 
we show plots for several sensor meta data from camera 2, namely, heading, accelerometer, gravity, gyroscope, magnetic, orientation, rotation and speed. 

\begin{figure*}[]
\begin{center}
  \includegraphics[width=1.0\linewidth]{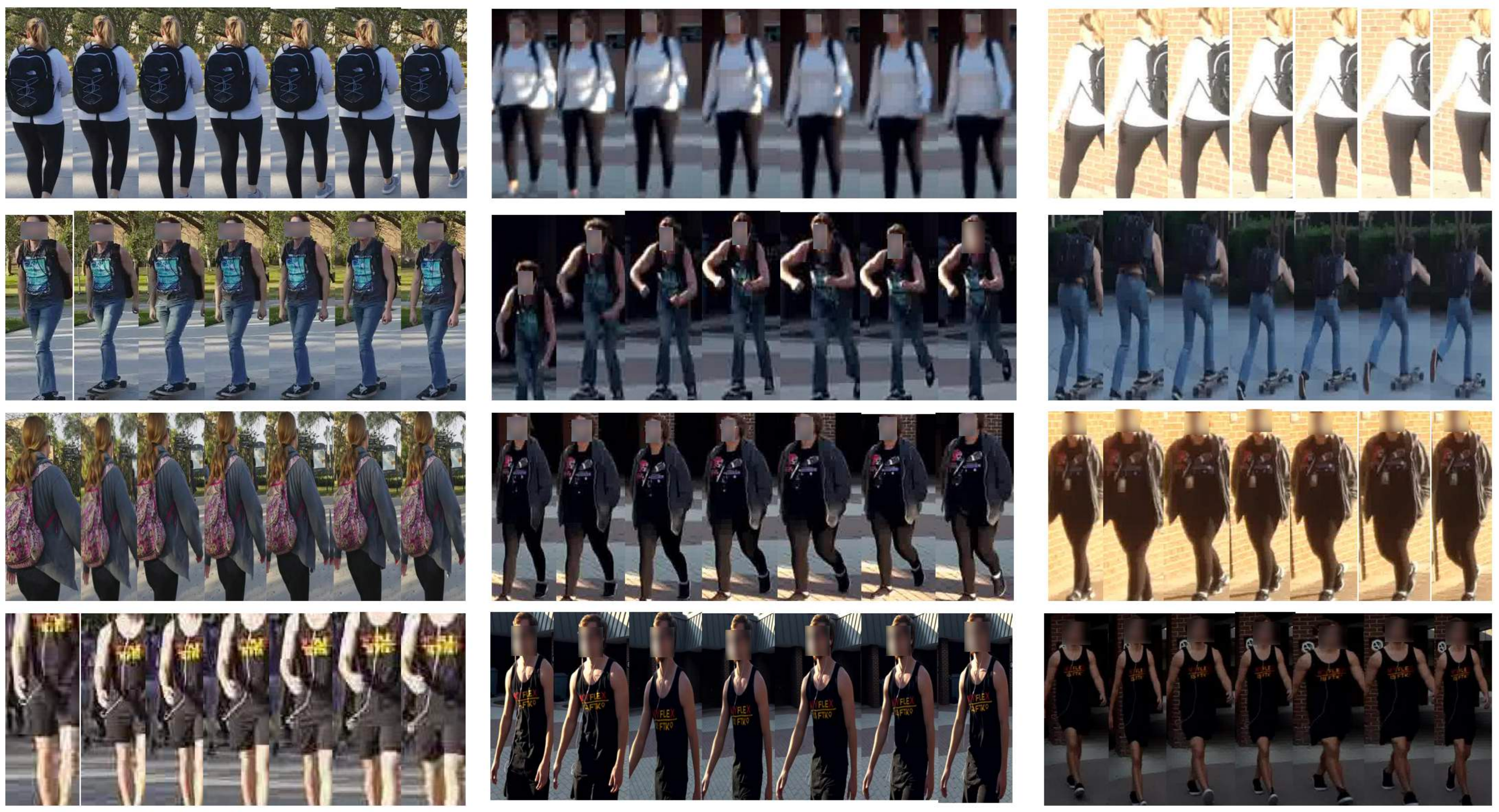}
\end{center}
  \caption{Sample tracklets from EgoReID. Each row corresponds to different identities, while each panel (from left to right) represents tracklets from cameras 1, 2 and 3.  As can be noted, in addition to changes in background, pose and illumination of the same tracklet across camera (each row), due to camera motion, different tracklets from the same camera (each panel) are also captured with different background, illumination and poses. This makes our dataset challenging but more close to the reality. In the above figure, for the sake of better visualization, we resize pedestrian detection boxes to the same size, but in our dataset they are of different sizes. 
  }
\label{fig:sample_Tracklet}
\end{figure*}

Fig. $\ref{fig:sample_Tracklet}$ shows sample tracklets from each camera. Each row contain tracklets of the same person, while each column represent different camera. As can be noted from fig. $\ref{fig:sample_Tracklet}$, our dataset contains several illumination, pose and background changes both within and across cameras.

\begin{figure}[t]
\begin{center}
  \includegraphics[width=1\linewidth]{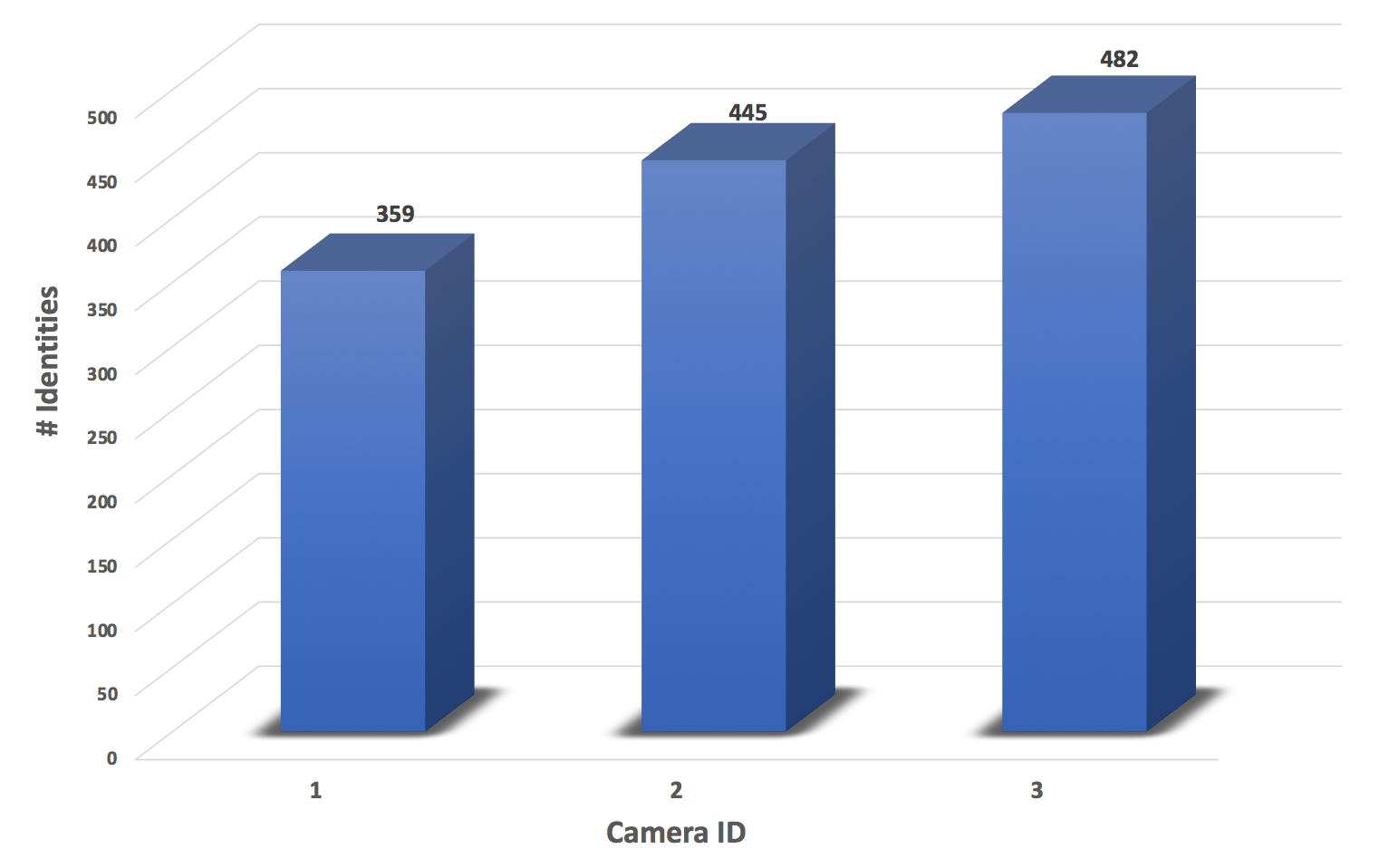}
\end{center}
  \caption{The number of IDs captured by each camera.}
\label{fig:long}
\label{fig:ID}
\end{figure}

\begin{figure}[t]
\begin{center}
  \includegraphics[width=1\linewidth]{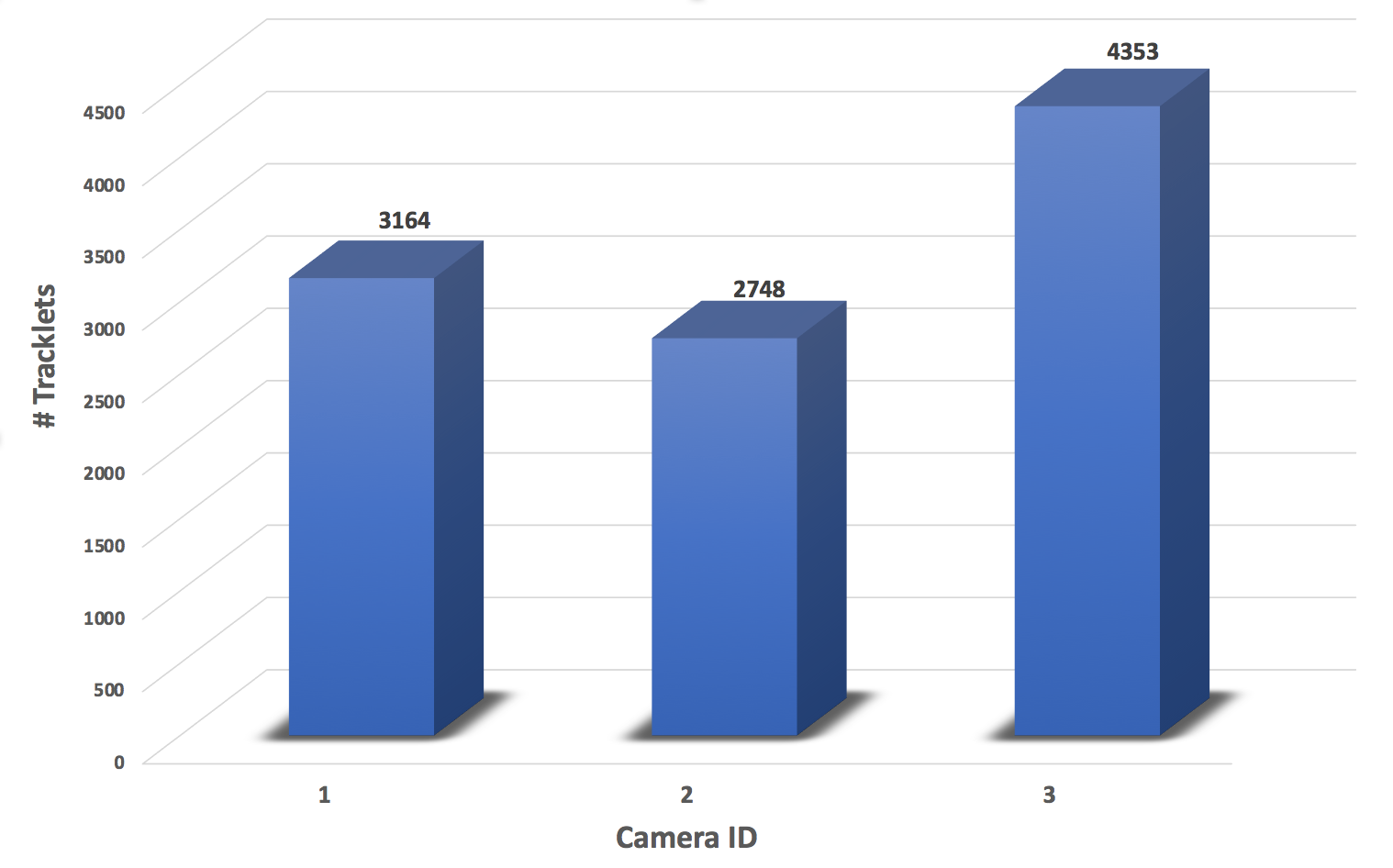}
\end{center}
  \caption{The number of tracklets per each camera.}
\label{fig:Tracklet}
\end{figure}

Fig. $\ref{fig:ID}$ and $\ref{fig:Tracklet}$ provide more detailed statistics on our dataset. Camera 3 captures the most number of IDs and tracklets. Camera 2 captures the second most number of IDs but ranked third in total number of tracklets it contains. 

Sample videos of detection and tracking results are included in the videos file. Moreover, we have included two video demonstrations for rotation, magnitude, orientation and gravity.



\section{Qualitative Results}
\begin{figure*}[t]
\begin{center}
\includegraphics[width=\textwidth]{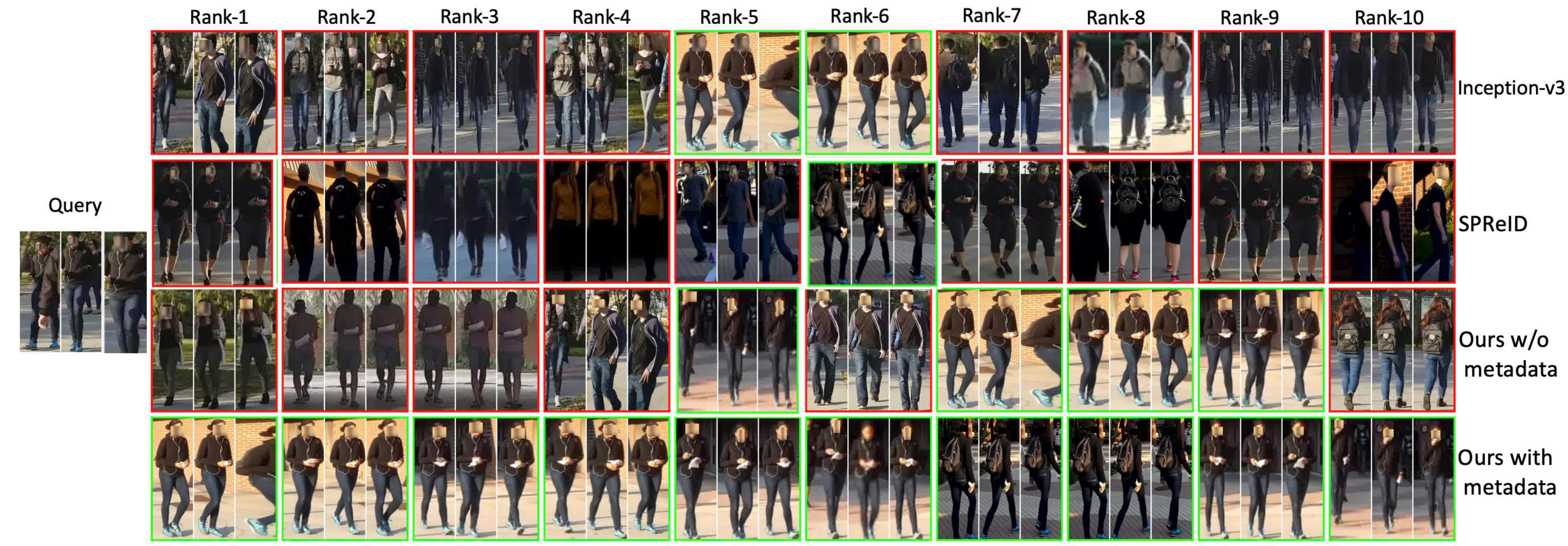}
\end{center}
  \caption{Inception-v3 and our approach without meta data are able to find the correct match among  top-4. However, after refining our results using sensor meta data, we are able to get the correct match at rank-1. }
\label{fig:s4}
\end{figure*}

\begin{figure*}[t]
\begin{center}
\includegraphics[width=\textwidth]{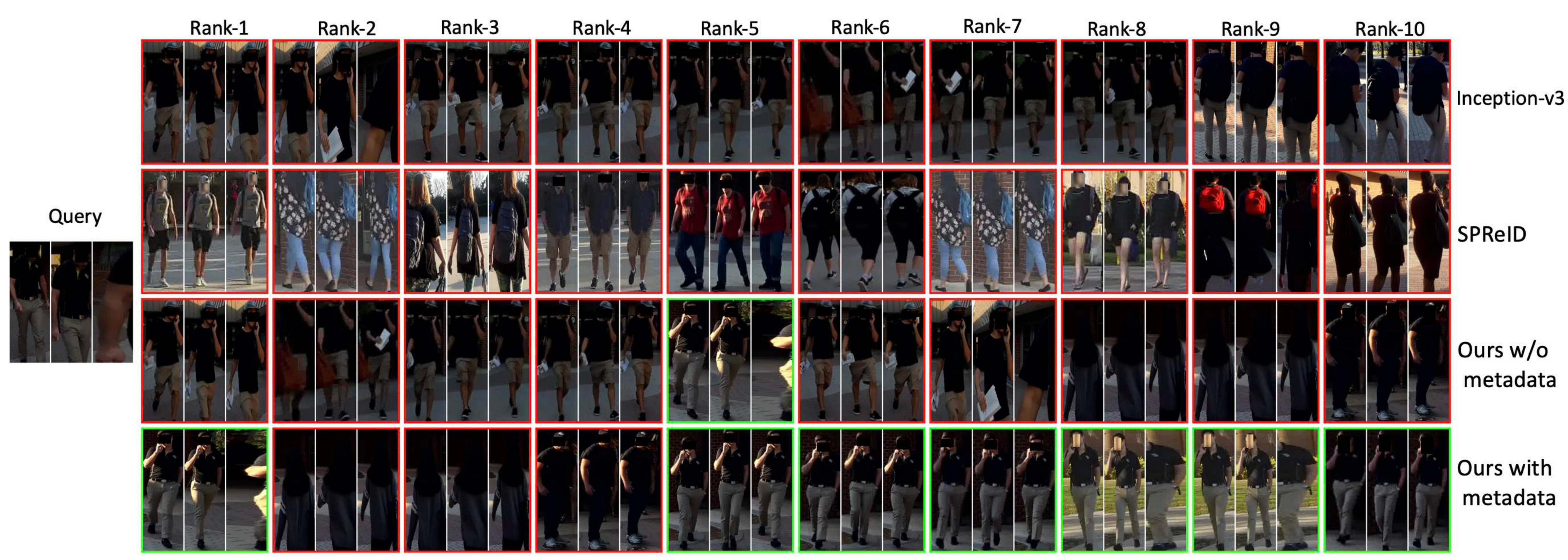}
\end{center}
  \caption{As it is evident that due to heavy occlusion of query tracklet, most approaches fail to find the correct match among  their top-10 results. Our method without using meta data finds the correct match at rank-5, while after applying meta data information the correct match is found at rank-1.}
\label{fig:s5}
\end{figure*}

\textbf{Person re-identification:}
In Fig. $\ref{fig:s4} - \ref{fig:s5}$, given different example queries from EgoReID dataset, top 10 retrieved tracklets from the gallery are shown for different methods. In all examples, the first row shows  the results obtained using Inception-v3, where we extract frame level global features and perform average pooling for each tracklet. The second row shows results, when we apply SPReID \cite{kalayeh2018human} on every frame in a tracklet and apply average pooling along the temporal dimension. While the third and fourth rows show results of our method before and after employing sensor meta data information. Correctly matched tracklets are shown with green boxes while the incorrect ones are show in red. To avoid clutter, we only show three frames per tracklet even though the length of tracklets in our dataset ranges between 16 and 31.

\textbf{Semantic Segmentation:}
To show the quality of our human semantic parsing model, we have provided videos which demonstrate the segmentation quality on randomly sampled tracklets from MARS \cite{zheng2016mars} and EgoReID datasets. In our videos, for better visualization, the original image, full body segmentation, and upper and lower body segmentations are shown side by side.

\section{Implementation Details}

\begin{figure}[t]
\begin{center}
  \includegraphics[height=7cm,width=7.6cm]{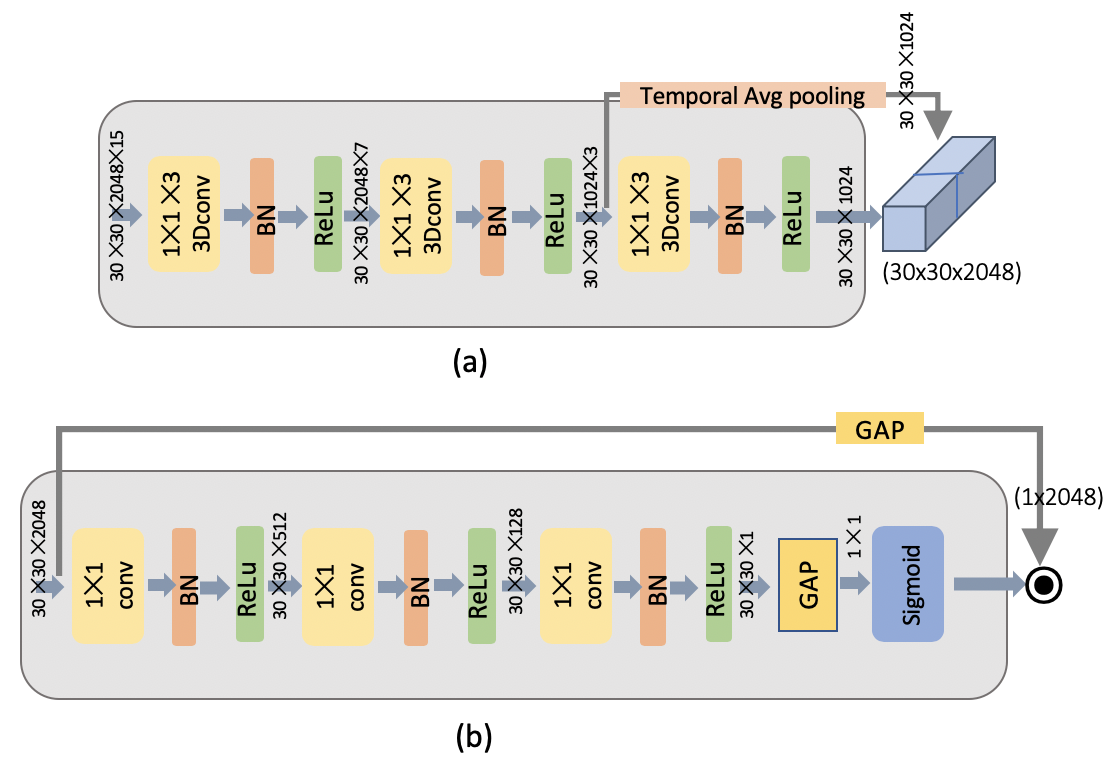}
\end{center}
  \caption{Detailed architecture used in our video feature extraction module. (a) shows one block of video feature extractor, the rest of the blocks have the same architecture. (b) shows details of a block from feature weighting module. We share weights between all blocks in feature weighting module.}
\label{fig:details}
\end{figure}

 Fig.\ref{fig:details} details the architecture of our video feature extraction module. Fig.\ref{fig:details}(a) shows one block of our video feature extraction module. All 3 block in this module have the same architecture. To exploit multi-scale information, after applying average pooling along the temporal dimension, we concatenate feature maps from intermediate layer with the output of the last layer.  Similarly, Fig.\ref{fig:details}(b) depicts the details of the proposed feature weighting module. We share parameters between our feature weighting modules. After applying global average pooling (GAP) on the output feature maps from video feature extraction module, we then use their corresponding weights to scale them.

\vspace{2mm}
\textbf{Parameters:} We train the Frame-level feature extraction module for 200K iterations using the training set consists of 10 image-based ReID datasets (Market-1501 \cite{zheng2015scalable}, CUHK01 \cite{li2012human}, CUHK02 \cite{li2013locally}, CUHK03 \cite{li2014deepreid}, DukeMTMC-reID \cite{zheng2017unlabeled}, 3DPeS \cite{baltieri20113dpes}, PRID \cite{hirzer2011person}, PSDB \cite{xiao2016end}, Shinpuhkan \cite{kawanishi2014shinpuhkan2014} and VIPeR \cite{gray2007evaluating}), then fine-tune it on the video datasets for 20K iterations. The initial learning rates for these two processes are 0.01 and 0.001, respectively, and we decay them 10 times with the rate of 0.9 by utilizing the exponential shift. Similarly,  we use 0.001 as the learning rate while training the video feature extraction module and decay it 10 times. While training the Frame-level feature extraction module, the batch size, momentum and weight decay are 15, 0.9 and 0.0005 for the respective values. To train the video feature extraction module, we set the batch size to 6 and the length of the clips to 15. The input images with the size of $512\times170$ are used to train both the frame-level feature extraction module and the video feature extraction module. 

We follow the same settings mentioned in \cite{kalayeh2018human} to train the semantic segmentation module. We set the initial learning rates to 0.01, 0.1 and 0.1 for the Inception-V3 backbone, atrous spatial pyramid pooling and the $1\times1$ convolution layer (classification layer), respectively, and decay them 10 times. The minibatch size is set to 8 and the input images with the size of $512\times512$ are used. The other settings are similar to the mentioned above. We perform the training of the re-identification and the segmentation models using Nesterov Accelarated Gradient \cite{bengio2013advances} and used the pre-trained InceptionV3 models on ImageNet \cite{russakovsky2015imagenet}.


\end{document}